\documentclass[journal,transmag]{IEEEtran}

\usepackage{mathtools}
\usepackage[export]{adjustbox}

\usepackage{lineno,hyperref}
\usepackage{multirow}
\usepackage{graphicx}
\usepackage{subfig}
\usepackage{enumitem}
\usepackage{setspace}
\usepackage{xcolor}

\modulolinenumbers[5]

\newcommand{\etal}{\textit{et al.}}
\newcommand{\tabincell}[2]{\begin{tabular}{@{}#1@{}}#2\end{tabular}}
\newcommand{\mf}[1]{\textcolor{black}{#1}}
\newcommand{\pr}[1]{\textcolor{black}{#1}}

\ifCLASSINFOpdf

\else

\fi

\begin{document}

\title{Real Masks and Spoof Faces: On the Masked Face Presentation Attack Detection}

\author{Meiling Fang$^{ a,b,*}$, Naser Damer$^{ a,b}$, Florian Kirchbuchner$^{ a}$, Arjan Kuijper$^{ a,b}$ \\
$^{a}$Fraunhofer Institute for Computer Graphics Research IGD, Darmstadt, Germany\\
$^{b}$Mathematical and Applied Visual Computing, TU Darmstadt,
Darmstadt, Germany\\
Email:meiling.fang@igd.fraunhofer.de
}

\maketitle

\begin{abstract}
Face masks have become one of the main methods for reducing the transmission of COVID-19. This makes face recognition (FR) a challenging task because masks hide several discriminative features of faces.
Moreover, face presentation attack detection (PAD) is crucial to ensure the security of FR systems. In contrast to the growing number of masked FR studies, the impact of face masked attacks on PAD has not been explored. Therefore, we present novel attacks with real face masks placed on presentations and attacks with subjects wearing masks to reflect the current real-world situation. Furthermore, this study investigates the effect of masked attacks on PAD performance by using seven state-of-the-art PAD algorithms under different experimental settings. We also evaluate the vulnerability of FR systems to masked attacks. The experiments show that real masked attacks pose a serious threat to the operation and security of FR systems.
\end{abstract}


\IEEEpeerreviewmaketitle

\section{Introduction}
Since the SARS-CoV-2 coronavirus outbreak and \pr{its} rapid global spread, wearing a mask has become one of the most efficient ways to protect and prevent \pr{getting infected with COVID-19}. However, \pr{for identity checks} in crowded scenarios \pr{such as at} airports, \pr{removing} the mask for face recognition (FR) increases \pr{the} chance \pr{of infection}. Wearing masks in public might be an essential health measure and a new norm even after the COVID-19 pandemic as most countries support the use of masks to minimize the spread of the virus.
As a result, researchers have shown an increased interest in the effect of face masks on the performance of FR verification \cite{ngan2020ongoing, DBLP:conf/biosig/DamerGCBKK20, IET_B_Mask}. \pr{The results of their studies have shown} that pre-COVID-19 FR algorithms suffer \mf{performance degradation} \pr{owing} to \pr{the} masked faces. 
However, \pr{attacks compromising the security and vulnerability of} FR systems for subjects wearing face masks have so far been overlooked. \pr{In this study,} security refer to the presentation attacks (PAs). Attackers can use PAs to spoof FR systems by impersonating someone or obfuscating their identity. Common PAs include printed photos/images, replayed videos \textcolor{black}{and 3D masks \cite{DBLP:journals/pr/JiaGX20,DBLP:journals/prl/JiaHLX21}}. Driven by the ongoing COVID-19 pandemic, presentation attack detection (PAD) \textcolor{black}{\cite{DBLP:journals/pami/YuWQLLZ21}} has encountered several understudied challenges when facing masked faces.
Current face PAD databases \cite{oulu_npu, Liu_auxiliary_siw_18, DBLP:conf/cvpr/LiuSJ019} only contain printed images or replayed videos \pr{in which} subjects were not wearing face masks. Therefore, there is uncertainty about the relationship between the performance of PAD techniques and PAs with face masks. Moreover, the vulnerability of FR systems \pr{to} masked attacks remains unclear. To \pr{overcome} such gaps, researchers require well-studied masked PAs.

\begin{figure}[htb!]
\begin{center}
\includegraphics[width=0.8\linewidth]{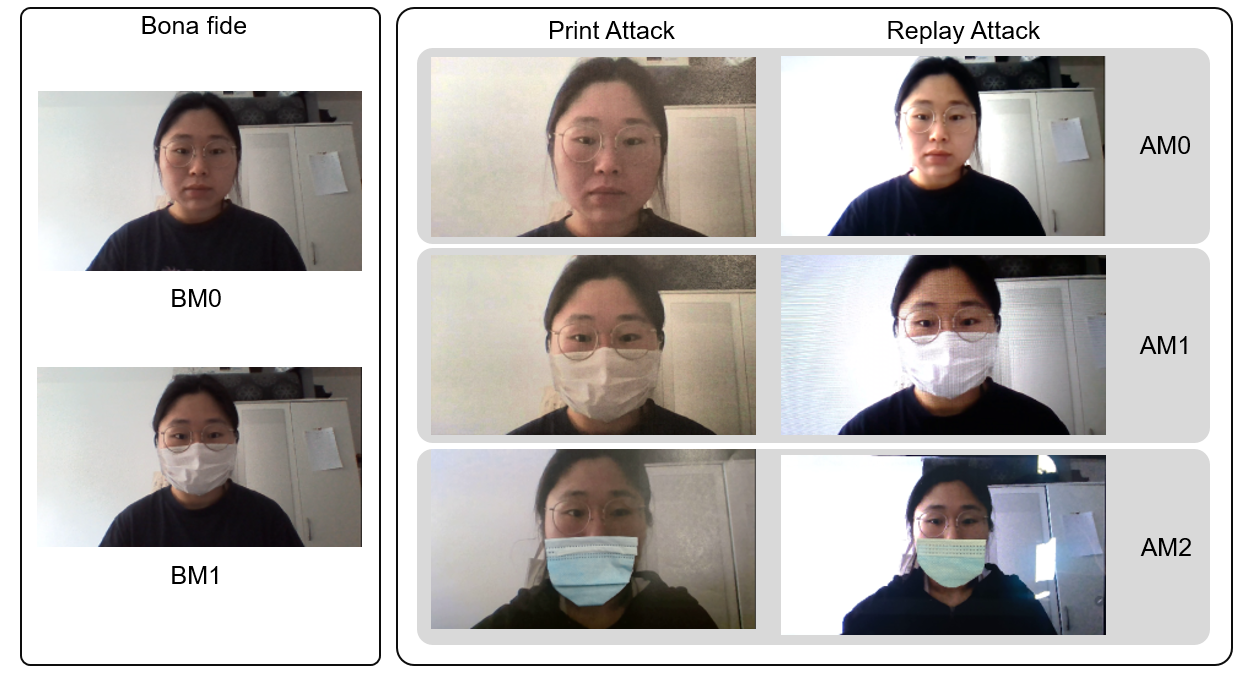}
\end{center}
\caption{Example bona fide and attack samples in \pr{the} CRMA database. \mf{Based on the presence of face masks, bona fides are grouped into BM0 (without mask) and BM1 (with mask) categories. The novel attacks are grouped into AM0 (spoof face without mask), AM1 (spoof face with mask), AM2 (spoof faces covered by real masks).}}
\vspace{-4mm}
\label{fig:image_examples}
\end{figure}

\mf{In this \pr{study}, we design and collect three types of attacks based on masked and unmasked face images collected realistically and collaboratively \cite{DBLP:conf/biosig/DamerGCBKK20, IET_B_Mask}. The bona fide samples \pr{were} divided into categories of BM0 (subjects wearing no masks) and BM1 (subjects wearing masks). AM0 data are unmasked print/replay attacks, \pr{which are} commonly \pr{used} data in most current PAD databases. AM1 data \pr{include} print/replay attacks, where  live subjects \pr{wore} masks. In addition, we provide a novel partial attack type, called AM2, where a real medical mask is placed on printed photos or replayed videos to simulate the subject wearing a mask. This is motivated by our assumption that AM2 might be a challenging attack as it contains both bona fide and attack presentations that may confuse PAD and/or FR systems. \pr{The} data samples are \pr{presented} in Fig.~\ref{fig:image_examples}.} 
The main contributions in this \pr{study} are:
\begin{itemize}[wide, labelwidth=!, labelindent=12pt]
\item The novel Collaborative Real Mask Attack Database (CRMA) is presented. Three types of \mf{PAs, called AM0 (unmasked face PA), AM1 (masked face PA), and AM2 (unmasked face PA with a real masked placed on the PA) (as shown in Fig.~\ref{fig:image_examples}), \pr{were} created for both print and replay presentation attack instruments (PAIs).}
\pr{To create} such attacks, three electronic tablets with high-resolution and three capture scales are used. Additionally, we \pr{designed} three experimental protocols \pr{to explore} the effect of masked attacks on PAD performance.
\item Extensive experiments are conducted to explore the effect of \mf{bona fide samples, masked faces attacks, and real masks (on spoof faces)} on the face PAD behavior. To support the comprehensive evaluation, seven PAD algorithms comprising of texture-based, deep-learning-based, and hybrid methods \pr{were} selected to evaluate the performance and generalizability in intra- and cross-database scenarios under three mask-related protocols. \pr{Both} quantitative and qualitative \pr{analyses revealed} that masked bona fides and PAs dramatically decreased the performance of PAD algorithms. Moreover, deep-learning-based methods perform worse on real mask attacks than mask-face attacks in most cases.  
\item An in-depth vulnerability analysis of FR systems is presented. We evaluated three deep-learning-based FR techniques \pr{for} three types of \mf{PAs}. The experimental results indicate that these three FR networks exhibit significantly higher vulnerabilities to the real mask attacks than masked face attacks.
\end{itemize}

We provide a brief review of relevant works in Sec.~\ref{sec:related_work}. \mf{Then, our novel CRMA database is described in detail in Sec.~\ref{sec:database}. The face PAD algorithms and FR systems \pr{used in this study} are introduced in Sec.~\ref{sec:experiments}. Sec.~\ref{sec:pad_results} introduces the three PAD protocols \pr{and} PAD evaluation metrics, and then discusses the PAD results. Sec.~\ref{sec:fr_results} describes the three FR experimental settings, used FR metrics, and analyzed the vulnerability of FR systems. Finally, \pr{conclusions are} presented in Sec.~\ref{sec:conclusion}.}

\vspace{-3mm}
\section{Related Work} 
\label{sec:related_work}
This section reviews the most relevant prior works from three perspectives: face PAD databases, face PAD methods, \mf{and FR and vulnerability analysis. At the end of each part, the difference between our work and prior \pr{work} is pointed out.}

\textbf{Face PAD Databases:} Data resources have become especially important \pr{ever} since \pr{the advent of deep learning}, because machine-learning-based algorithms have the risk of underfitting or overfitting on limited data. Given the significance of good-quality databases, several face PAD databases \pr{have been} released, \pr{such as} NUAA \cite{DBLP:conf/eccv/TanLLJ10}, CASIA-FAS \cite{casia_fas}, Replay-Attack \cite{replay_attack}, MSU-MFSD \cite{msu_mfs}, OULU-NPU \cite{oulu_npu}, and SiW \cite{Liu_auxiliary_siw_18}, all consisting of 2D print/replay attacks. In addition, SiW-M \cite{DBLP:conf/cvpr/LiuSJ019} and \mf{CelebA}-Spoof \cite{celea-spoof} databases provide multiple types of attacks \pr{such as} makeup, 3D mask, or paper cut. Moreover, some multimodal databases are publicly available: 3DMAD \cite{3dmad}, Mssproof \cite{msspoof-2015}, CASIA-SURF \cite{DBLP:journals/tbbis/ZhangLWLGEEL20}, \pr{and} CSMAD \cite{DBLP:conf/btas/BhattacharjeeMM18}.

These databases \pr{undoubtedly} contribute to the significant progress of PAD research. \pr{For example}, \mf{the} CeleA-Spoof database \pr{comprises} images from various environments and illuminations with rich annotations to reflect real scenes. However, these databases also have weaknesses: 1) the multimodal databases have high hardware requirements and cannot be widely used in daily life; 2) some databases \pr{such as} CASIA-MFS \cite{casia_fas} and MSU-MFS \cite{msu_mfs} cannot satisfy the current needs because of the lower quality of the outdated acquisition sensors; 3) Oulu-NPU \cite{oulu_npu}, SiW \cite{Liu_auxiliary_siw_18}, SiW-M \cite{DBLP:conf/cvpr/LiuSJ019}, and CelebA-Spoof \cite{celea-spoof} are relatively up-to-date, but \pr{they do not consider} PAs with real face masks to fit the current COVID-19 pandemic. Hence, we collect the CRMA database to fill the gaps \pr{in} these databases \pr{in the context of} the ongoing COVID-19 pandemic; \pr{furthermore, we} ensure the \pr{the database is} generalizable and compatible with real scenarios. The CRMA database can be used to better analyze the effect of a real mask on PAD performance and the vulnerability of FR systems for novel attacks, such as placing a real mask on an attack presentation. Detailed information related to the databases \pr{mentioned above} is \pr{presented} in Tab.~\ref{tab:db_summary}).

\begin{table*}[thbp!]
\centering
\def\arraystretch{1.4}
\resizebox{\textwidth}{!}{%
\begin{tabular}{|c|c|c|c|c|c|c|c|}
\hline
Database & Year & \# Subjects & \# Data (BF/attack) & Capture devices (BF/attack) & Display devices & Modality & Attack type \\ \hline
NUAA \cite{DBLP:conf/eccv/TanLLJ10} & 2010 & 15 & 5105/7509 (I) & Webcame & - & RGB & 1 Print \\ \hline
CASIA-FAS \cite{casia_fas} & 2012 & 50 & 150/450 (V) & Two USB cameras, Sony NEX-5 & iPad & RGB & 1 Print, 1 Replay  \\ \hline
Replay-Attack \cite{replay_attack} & 2012 & 50 & 200/100 (V) & MacBook 13 / iPhone 3GS, Cannon SX150 & iPhone 3GS, iPad & RGB & 1 Print, 2 Replay \\ \hline
3DMAD \cite{3dmad} & 2013 & 17 & 170/85 (V) & Microsoft Kinect & - & RGB/Depth & 1 3D Mask\\ \hline 
Msspoof \cite{msspoof-2015} & 2015 & 21 & 1,680/3,024 (I) & uEye camera & - & RGB/IR & 1 Print \\ \hline
MSU-MFSD \cite{msu_mfs} & 2015 & 35 & 110/330 (V) &  \tabincell{c}{ MacBook Air, Google Nexus 5 / \\ Cannon 550D, iPhone 5s} & iPad Air, iPhone 5s & RGB & 1 Print, 2 Replay \\ \hline
Oulu-NPU \cite{oulu_npu} & 2017 & 55 & 1,980/3,960 (V) & 6 smartphones & Dell 1905FP, Macbook Retina & RGB & 2 Print, 2 Replay  \\ \hline
SiW \cite{Liu_auxiliary_siw_18} & 2018 & 165 & 1,320/3,300 (V) & Cannon EOS T6, Logitech C920 webcam & \tabincell{c}{iPad Pro, iPhone 7, \\ Galaxy S8, Asus MB 168 B} & RGB & 2 Print, 4 Replay  \\ \hline 
CASIA-SURF \cite{DBLP:journals/tbbis/ZhangLWLGEEL20} & 2018 & 1000 & 18000/3000 (I) & RealSense camera & - & RGB/IR/Depth & 5 Papercut \\ \hline
CSMAD \cite{DBLP:conf/btas/BhattacharjeeMM18} & 2018 & 14 & 88/160 (V) & RealSense, Compact Pro, Nikon P520 & - & RGB/IR/Depth/LWIR & 1 silicone mask \\ \hline
SiW-M \cite{DBLP:conf/cvpr/LiuSJ019} & 2019 & 493 &  660/1630 (V) & Logitech C920, Cannon EOS T6 & - & RGB & \tabincell{c}{ 1 Print, 1 Replay, \\ 5 3D mask, 3 Makeup, 3 Partial} \\ \hline
Celeb-Spoof \cite{celea-spoof} & 2020 & 10,177 & 202,559/475,408 (I) & \tabincell{c}{ Various cameras/ 20 smartphones, \\ 2 webcams, 2 tablets} & PC, phones, tablets, & RGB & \tabincell{c}{ 3 Print, 3 Replay, \\ 1 3D mask, 3 Paper Cut} \\ \hline \hline
CRMA & 2021 & 47 & 423/12,690 (V) & \tabincell{c}{ Webcams/iPad Pro, \\ Galaxy Tab S6, Surface Pro 6} & \tabincell{c}{iPad Pro, Galaxy Tab S6, \\ Surface Pro 6} & RGB & \tabincell{c}{1 Print, 3 Replay, \\ \textbf{1 Real mask} } \\ \hline
\end{tabular}}
\caption{The summary of face PAD databases, including our CRMA database information for brief comparison. It should be noted that our CRMA database is the only database containing subjects wearing face masks and real face mask attacks. The details of our CRMA database \pr{are presented} in Sec.~\ref{sec:database}.}
\vspace{-4mm}
\label{tab:db_summary}
\end{table*}

\textbf{Face PAD Methods:} In recent years, there has been an increasing \pr{number} of studies in the \pr{field of} face PAD \textcolor{black}{\cite{DBLP:journals/pr/SongZFL19,DBLP:journals/pr/JiaZSC21,DBLP:journals/pr/FatemifarAAK21}}. These studies can be broadly grouped into three categories: texture-based methods, deep-learning-based methods, and hybrid methods. Texture features, such as local binary pattern (LBP) \cite{DBLP:journals/pami/OjalaPM02}, project the faces to a low-dimensional embeddings. \mf{M{\"{a}}{\"{a}}tt{\"{a}}} \etal \cite{pad_lbp_2011} proposed an approach using multi-scale LBP to encode the micro-texture patterns into an enhanced feature histogram for face PAD. The resulting histograms were then fed to a support vector machine (SVM) classifier to determine whether a sample is \pr{a} bona fide or attack. The LBP features extracted from different color spaces \cite{DBLP:conf/icip/BoulkenafetKH15} were further proposed to utilize chrominance information. They achieved competitive results on Replay-Attack \cite{replay_attack} (equal error rate (EER) value of 0.4\%) and CASIA-FAS \cite{casia_fas} (EER value of 6.2\%) databases. Furthermore, Boulkenafet \etal \cite{pad_competition} organized a face PAD competition based on the OULU-NPU database and compared 13 algorithms provided by participating teams and one color-LBP-based method \mf{(referred to as baseline in \cite{pad_competition})}. In this competition, the GRADIANT algorithm \mf{fused multiple information, \pr{that is}, color, texture, and motion. The GRADIANT} achieved competitive results in \pr{the} four evaluation protocols. In addition to the \mf{texture-based} GRADIANT approach, deep-learning-based method (MixFASNet) or hybrid method (CPqD) also achieved lower error rates in all experimental protocols. CPqD fused the results from \pr{the} fine-tuned Inception-v3 network and the \mf{color-LBP-based method (referred to as \pr{the} baseline in \cite{pad_competition}).} Consequently, we chose \mf{to} re-implement \pr{the} color-LBP and CPqD methods in this \pr{study} (details in Sec.~\ref{ssec:pad_algorithms}), while the GRADIANT and MixedFASNet are discarded in our work \pr{because} they do not provide \pr{sufficient} details for re-implementation. Deep-learning-based methods have been pushing the frontier of face PAD research and have shown remarkable improvements in PAD performance. 
Lucena \etal \cite{FASNet} presented an approach \pr{called} FASNet \pr{in which} a pre-trained VGG16 is fine-tuned by replacing the last \pr{fully connected} layer. The FASNet network achieved \pr{excellent} performance on 3DMAD \cite{3dmad} and Replay-Attack databases \cite{replay_attack}. Recently, George \etal \cite{deeppix_19} proposed \mf{training a network with pixel-wise binary supervision on feature maps to exploit information from different patches.}  DeepPixBis \cite{deeppix_19} outperformed \pr{the} state-of-the-art algorithms in Protocol-1 of \pr{the} OULU-NPU database (0.42\% ACER) but also achieved \pr{significantly} better results than traditional texture-based approaches in the cross-database scenario. Considering the popularity of PAD techniques and the ease of implementation, we also chose FASNet and DeepPixBis (details in Sec.~\ref{ssec:pad_algorithms}) to study the effect of the real mask and masked face attacks on \pr{the} PAD performance.  

\textbf{Face Recognition and Vulnerability Analysis:} As one of the most popular modalities, the face has received increasing attention in authentication/security processes, such as smartphone face unlocking and automatic border control \mf{(ABC)}. Moreover, FR techniques \cite{arcface, sphereface, vggface2} have achieved significant performance improvements, and many personal electronic products have deployed FR technology. 
However, the ongoing COVID-19 pandemic brings a new challenge related to the behavior of collaborative recognition techniques when dealing with masked faces. \mf{Collaborative data collection refers to a subject actively attending to use the FR systems, such as unlocking personal devices or using an ABC gate, in contrast to uncollaborative capture scenario where the user does not intentionally use the FR service, \pr{such as in the case of} surveillance.}
\pr{The} National Institute of Standards and Technology (NIST) \cite{ngan2020ongoing} provided a preliminary study that evaluated the performance of 89 commercial FR algorithms developed before the COVID-19 pandemic. Their results indicated that digitally applied face masks with photos decreased the recognition accuracy; \pr{for example}, even the best of the 89 algorithms had error rates between 5\% \pr{and} 50\%. It is worth noting that the masks used in \pr{the} experiments were synthetically created. Damer \etal \cite{DBLP:conf/biosig/DamerGCBKK20, IET_B_Mask} presented a real mask database to simulate a realistically variant collaborative face capture scenario. \mf{Each participant was asked to simulate a login scenario by actively looking toward a capture device, such as a static webcam or a mobile phone. Our attack samples were created and collected based on the masked face data, which refers to the bona fide samples in the PAD case (as described in Sec.~\ref{sec:database}).} They also explored the effect of wearing a mask on FR performance and concluded that face masks significantly reduce the accuracy of algorithms. 
Mohammadi \etal \cite{DBLP:journals/iet-bmt/MohammadiBM18} \pr{provided} empirical evidence to support the claim that the CNN‐based FR methods are extremely vulnerable to 2D PAs. Subsequently, Bhattacharjee \etal \cite{DBLP:conf/btas/BhattacharjeeMM18} presented the first FR-vulnerability study on 3D PAs. The experiments also clearly showed that CNN-based FR methods are vulnerable to custom 3D mask PAs. However, the vulnerability of FR systems \pr{to} \mf{PAs with face masks} has not been investigated. Therefore, in this \pr{study}, we selected three CNN-based FR algorithms for further FR-vulnerability analysis on masked face attacks: the state-of-the-art ArcFace \cite{arcface}, SphereFace \cite{sphereface}, and VGGFace \cite{vggface2}. These algorithms are discussed in more detail in Sec.~\ref{ssec:recognition_algo}.

\vspace{-3mm}
\section{The Collaborative Real Mask Attack Database (CRMA)} 
\label{sec:database}

Our proposed CRMA database \footnote{The CRMA database is not publicly available \pr{because of} privacy regulations. However, the database will be (1) available for assisted in-house research use by collaborators and partners in the research community; (2) bending the legal authorization by the data collection institute, the data will be submitted to be included on the Open Science BEAT platform (www.beat-eu.org).} \pr{and} can serve as a supplement to the databases in Tab.~\ref{tab:db_summary}, and \pr{because of} the COVID-19 pandemic, it can better reflect \pr{the} possible issues facing real-world PAD performance. \mf{The CRMA database includes 1) both unmasked (BM0) and masked (BM1) bona fide samples collected in a realistic scenario \cite{DBLP:conf/biosig/DamerGCBKK20,IET_B_Mask}, 2) conventional replay and print PAs created from faces not wearing a mask (AM0), 3) replay and printed PAs created from masked face images (AM1), and 4) novel PAs where the PAs of unmasked faces are covered (partially) with real masks (AM2), as shown in Fig.~\ref{fig:image_examples}.
Damer et al. \cite{DBLP:conf/biosig/DamerGCBKK20, IET_B_Mask} collected data to investigate the effect of wearing a mask on face verification performance. For PAD, such data \pr{are} considered bona fide. The data presented in this \pr{study} build on an extended version of the data introduced in \cite{DBLP:conf/biosig/DamerGCBKK20,IET_B_Mask}, by creating and capturing different \pr{types} of PAs based on the bona fide data captured in \cite{DBLP:conf/biosig/DamerGCBKK20, IET_B_Mask}. As a result, the bona fide data in this work \pr{are} an extended version of the one introduced in \cite{DBLP:conf/biosig/DamerGCBKK20, IET_B_Mask} and the attack data presented here \pr{are} completely novel and \pr{have} not \pr{been} previously studied.}

Fig.~\ref{fig:statistics_info} introduces the general statistical information of the CRMA database. This database contains 62\% males and 38\% females. The attack \mf{AM0, AM1, and AM2} ratios are 30\%, 60\%, and 10\%, respectively, as will be described later in this section. Additionally, we count the frequency of the proportion of the face size in the video. The histogram shows that the proportion of the face areas in the videos is mostly between 5\% and 30\%. 
This section \mf{first describes the bona fide samples provided by \cite{DBLP:conf/biosig/DamerGCBKK20, IET_B_Mask}, and then introduces our process of attack sample creation and collection.} 

\subsection{Collection of bona fide samples}
\pr{To explore the FR performance on masked faces,} Damer \etal \cite{DBLP:conf/biosig/DamerGCBKK20, IET_B_Mask} recently presented a database \mf{where} the subjects wearing face masks.  

This database simulates a collaborative \mf{environment in which participants collect videos by actively looking towards the capture device. During this process, the eyeglasses were removed when the frame \pr{was} considered very thick following the International Civil Aviation Organization (ICAO) standard \cite{icao}. The videos were captured by the participants at their residences while working from home. Therefore, the types of face masks, capture devices, illumination, and background were \pr{varied}. For PAD, these videos are classified as bona fides and will be used later to create attack samples.} 

\mf{The final version \cite{IET_B_Mask} of this database contains 47 participants. Each subject recorded a total of nine videos over three days with three different scenarios for each day. In contrast to the study by Damer \etal \cite{DBLP:conf/biosig/DamerGCBKK20}, which examined the effects of both face masks and illumination variations, we focus\pr{ed} only on the impact of face masks on PAD performance.
Hence, in our study, \pr{the} bona fide videos are} divided into two categories: a face without a mask on is \pr{denoted} as \mf{BM0} (three videos per subject), and a face with a mask is marked as \mf{BM1} (six videos per subject) \mf{(as shown in the right column of  Fig.~\ref{fig:image_examples})}.


\subsection{Creation of the presentation attacks}
\begin{figure}[htb!]
\begin{center}
\includegraphics[width=0.99\linewidth]{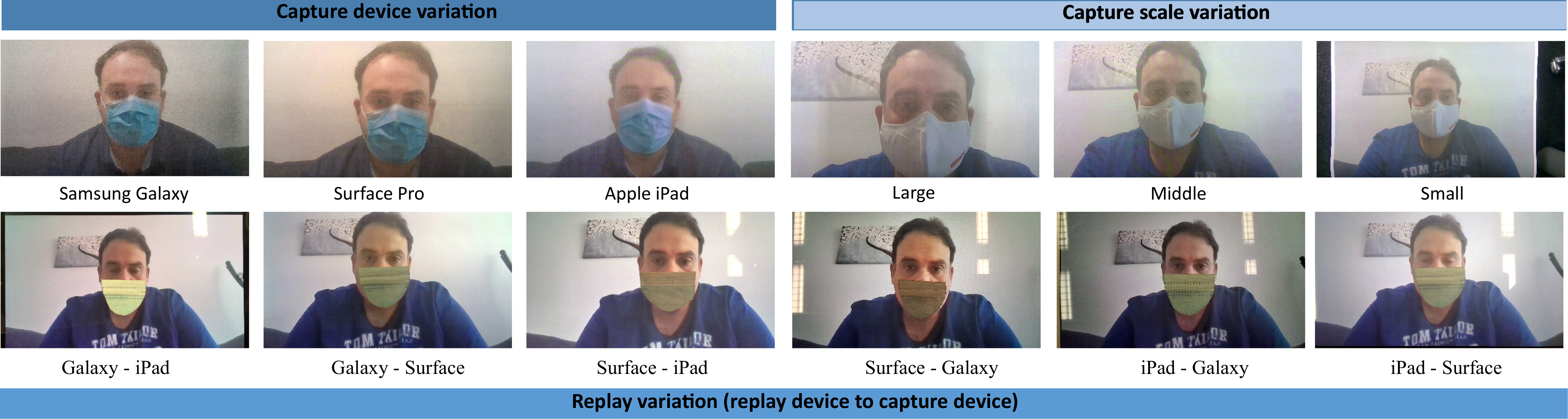}
\end{center}
\caption{Different capture variations in the CRMA database. The top left \pr{shows the} videos captured by different devices. The top right \pr{shows} the different capture scales. The bottom \pr{shows} the six cross-device types of replay attack settings.}
\label{fig:capture_variation}
\end{figure}

\begin{figure}[htb!]
\begin{center}
\includegraphics[width=0.99\linewidth]{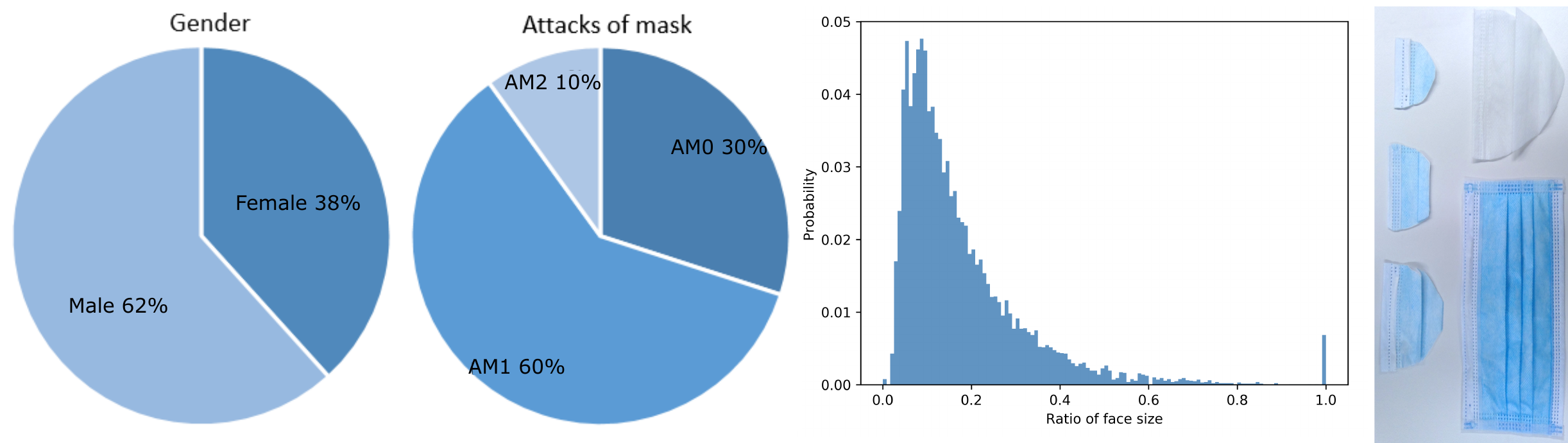}
\end{center}
\caption{The statistics of the subjects and the used mask shapes for creating \mf{AM2} samples in the CRMA database. From left to right: gender, mask types of attacks (\mf{AM0, AM1, AM2}), the histogram shows the probability distribution of the face size ratio and the applied mask shapes.}
\vspace{-3mm}
\label{fig:statistics_info}
\end{figure}

\mf{Most} FR databases tried to collect data under various harsh conditions, such as poor lighting, strong occlusion, or low resolution. Such databases \pr{attempted} to reproduce what might happen in a real-world scenario when a legitimate user obtains authorization \cite{LFWTech}. \pr{In contrast}, attackers use highly sophisticated artifacts, such as high-resolution images or videos, to maximize the success rate when \mf{impersonating} someone. 
For \pr{this} reason, \mf{we first collect the PAs} in a windowless room where all lights \pr{are} on. \mf{Second}, three high-resolution electronic tablets were used in the acquisition process: 1) iPad Pro (10.5-inch) with the display resolution of $ 2224 \times 1668$ pixels, 2) Samsung Galaxy Tab S6 with the display resolution of $2560 \times 1600$ pixels, 3) Microsoft Surface Pro 6 with the display resolution of $2736 \times 1824$ pixels. \mf{In the process of collecting data}, the capture devices and \mf{displayed} images/tablets were stationary. The videos were captured with \pr{a resolution of} $1920 \times 1080$. \pr{In addition}, each video had a minimum length of 5 seconds, and the frame rate \pr{was} 30 fps. \mf{This work focuses on the two common PAIs, print and replay attacks, due to their ease of creation and low cost.} The attack data in each PAI (\pr{see the} samples in Fig.~\ref{fig:image_examples}) are divided into three types: 1) \mf{the spoof face with no face mask (AM0), 2) the spoof face with a mask on (AM1), and 3) the spoof face with no face mask, but a real mask was placed on it to simulate a participant wearing a mask (AM2)}. However, the \mf{size of the} face area \mf{in each video} is slightly inconsistent because the videos were recorded by the participants themselves. To reproduce the \pr{appearance} of wearing a mask in the real world, we cropped five face masks to fit most of the faces (see Fig. ~\ref{fig:statistics_info}). \mf{The five masks are three small blue surgical masks, one slighter bigger white face mask, and one uncropped mask. When placing the mask, we select a suitable mask according to the size of the face in the printed image or video, aiming to cover the nose to the chin area and the cheeks without exceeding.} The details of each PAI \pr{are} as followings:\\
\textbf{Print image attack}: In print \mf{PAI}, an attacker tries to fool the FR system using a printed photo. 
\mf{Considering the instability of the face during the first second, such as the participant pressing the recording button or adjusting the sitting position, the 35\textsuperscript{th} frame of each bona fide video was printed out as an attack artifact.}  
Therefore, we obtain\pr{ed} nine photos per subject. \pr{The three tablets mentioned above} were used to capture the photos. Furthermore, to increase the diversity and variety of the data, each tablet captured three videos for a photo with three scales (see examples in Fig.~\ref{fig:capture_variation}). The captured videos using the first scale contain\pr{ed} all areas (100\%) of the photos, the second scale consist\pr{ed} of most areas (80\%) of the original photos, and the third scale focus\pr{ed} on the face area (60\%) as much as possible.  In addition to collecting attack data \pr{solely} from printed images, we also collected data from real face masks overlaid on photos (i.e., the previously defined \mf{AM2}). Theoretically, real masks will reduce the region of artificial features and increase the complexity and mixture of the features in the collected attack data. Eventually, 90 print attack videos \pr{were} generated for each subject, \pr{that is}, a total of 4,230 videos for 47 subjects in print PAI. \\
\textbf{Replay video attack}: In replay \mf{PAI}, an attacker tries to obtain the authentication by replaying a video. The three common points of the collection process between print and replay PAI are the use of three tablets, the use of three scales, and the \mf{process of AM2 data creation}, respectively. The difference is that these tablets were \pr{also} used for capturing displays of videos (see examples in Fig.\ref{fig:capture_variation}). While one tablet was replaying the video, the other two tablets were used to capture the data. As a result, each subject corresponded to 180 replay attack videos (162 videos of the \mf{AM0 and AM1} groups, 18 videos of \mf{AM2}.), i.e., \pr{there were} a total of 8,460 videos in this attack subset.

\section{\mf{Experimental Algorithms}} 
\label{sec:experiments}
This section first describes the adopted face PAD algorithms for the investigation of masked face attacks. \pr{Subsequently}, three FR algorithms were introduced for further vulnerability analysis. In both PAD and FR experiments, the widely used multi-task cascaded convolutional networks (MTCNN) \cite{DBLP:journals/spl/ZhangZLQ16} technique \pr{was} adopted to detect and crop the face.

\subsection{Face PAD algorithms}
\label{ssec:pad_algorithms}
A competition \cite{pad_competition} was carried out in 2017 to evaluate and compare the generalization performance of face PAD techniques under real-world variations. In this competition \cite{pad_competition}, there were 14 participating teams \mf{together with} organizers that contributed \pr{to} several state-of-the-art approaches. We chose two methods ((as previously discussed in Sec.~\ref{sec:related_work})), \pr{the} LBP-based method (referred to as \pr{the} baseline in \cite{pad_competition}), and \mf{hybrid} CPqD, and included additional solutions. We re-implemented a total of seven face PAD algorithms in this study, which can be categorized into three groups: hand-crafted features, deep-learning features, \pr{and} hybrid features. For further cross-database evaluation scenarios, we used three publicly available databases, mainly involving 2D PAs (details in Sec.~\ref{sec:related_work}): CASIA-FAS \cite{casia_fas}, MSU-MFS \cite{msu_mfs}, and OULU-NPU \cite{oulu_npu} in the competition. A brief description of the adopted methods \pr{is} provided below: 
\begin{itemize}[wide, labelwidth=!, labelindent=12pt]
    \item \textbf{LBP:} The LBP \mf{method is referred to as} baseline method in \cite{pad_competition} provided by the competition organizers that \mf{utilized} the color texture technique. The face in a frame is first detected, cropped, and normalized \pr{to} a size of $64 \times 64$ pixels. Second, an RGB face \pr{was} converted into HSV and YCbCr color spaces. Third, the LBP features \pr{were} extracted from each channel. The obtained six LBP features are then concatenated into one feature vector to feed into a softmax classifier. The final prediction score for each video \pr{was} computed by averaging the output scores of all the frames.
    \item \textbf{CPqD:} The CPqD is based on the Inception-v3 network \cite{inception_v3} and the above LBP method. The last layer of the pre-trained Inception-v3 model \pr{was} replaced by a fully connected layer and a sigmoid activation function. The faces in the RGB frames are detected, cropped, and normalized \pr{to} $299 \times 299$ pixels. These face images \pr{were} utilized as inputs to fine-tune the Inception-v3 model. The model with the lowest EER on the development set among all 10 training epochs \pr{was} selected. A single score for each video \pr{was} obtained by averaging the output scores of all frames. To further improve the performance, the final score for each video \pr{was} computed by fusing the score achieved by the Inception-v3 model and the score obtained by the LBP method.
    \item $\mathrm{\mathbf{Inception_{FT}}}$ and $\mathrm{\mathbf{Inception_{TFS}}}$: Since the CPqD uses the Inception-v3 \cite{inception_v3} network as the basic architecture, we also report the results of fine-tuned Inception-v3 model, named $\mathrm{Inception_{FT}}$. In addition to the \mf{fine-tuned} model, we \pr{trained} the Inception-v3 model from scratch for performance comparison, named $\mathrm{Inception_{TFS}}$. In the training phase, the binary cross-entropy loss function and Adam optimizer with a learning rate of $10^{-5}$ \pr{were} used. The output scores of the frames \pr{were} averaged to obtain a final prediction decision for each video.
    \item $\mathrm{\mathbf{FASNet_{FT}}}$ and $\mathrm{\mathbf{FASNet_{TFS}}}$: FASNet \cite{FASNet} used transfer learning from pre-trained VGG16 model \cite{vgg16} for face PAD. They used a pre-trained VGG16 model as a feature extractor and modified the last fully connected layer. The newly added fully connected layers with a sigmoid function were then fine-tuned for the PAD task. This fine-tuned FASNet is referred to \pr{as} $\mathrm{FASNet_{FT}}$, similar to the Inception-v3 network methods, \pr{and} we also train FASNet from scratch with the name $\mathrm{FASNet_{TFS}}$. The input images are the detected, cropped, and normalized RGB face frames with \pr{a} size of $224 \times 224$ pixels. The Adam optimizer with \pr{a} learning rate of $10^{-4}$ \pr{was} used for training, as defined in \cite{FASNet}. Data augmentation techniques and class weights are utilized \pr{to deal with imbalanced data problems.}. To further reduce overfitting, \mf{an} early stop technique with a patience of 5 and maximum epochs of 30 was used. The resulting scores were averaged to obtain the final score for each video.
    \item \textbf{DeepPixBis:} George \etal \cite{deeppix_19} proposed a densely connected network framework for face PAD with binary and deep pixel-wise supervision. This framework is based on DenseNet \pr{architecture} \cite{densenet}. Two dense blocks and two transition blocks with a fully connected layer with sigmoid activation produce \pr{a} binary output. We \pr{used} the same data augmentation technique (horizontal flip,  random \mf{jitter} in brightness, contrast, and saturation) and the same hyper-parameters (Adam optimizer with a learning rate of $10^{-4}$ and weight decay of $10^{-5}$) as defined in \cite{deeppix_19} for the training. In addition to data augmentation, we \pr{applied the} class weight and \mf{an} early stopping technique to avoid overfitting. The final score for each video \pr{was} computed by averaging the frame scores.
\end{itemize}

\subsection{Face Recognition algorithms}
\label{ssec:recognition_algo}
For FR systems, trained CNNs are typically used as feature extractors. The feature vector extracted from a specific layer of an off-the-shelf CNN \pr{was} used as the template to represent the corresponding input face image. Then, the resulting templates \pr{were} compared with each other using similarity measures. To provide a vulnerability analysis of the FR systems to our novel masked attacks, we \pr{adapted} the following three FR algorithms:
\begin{itemize}[wide, labelwidth=!, labelindent=12pt]
    \item \textbf{ArcFace:} ArcFace \cite{arcface} introduced an additive angular margin loss function to obtain highly discriminative features for FR. We \pr{chose} this algorithm because ArcFace consistently outperformed state-of-the-art \pr{methods}. ArcFace achieved 99.83\% on Labeled Faces in the Wild (LFW) \cite{LFWTech} and 98.02\% on YouTube Faces (YTF) \cite{YTF} dataset. The pre-trained ArcFace model \footnote{The official ArcFace model: \url{https://github.com/deepinsight/insightface}} in our study was based on the ResNet-100 \cite{resnet} architecture and trained on the MS-Celeb-1M \cite{ms1m} dataset (MS1M-v2). The output template is a 512-dimension feature vector extracted from the '\textit{fc1}' layer of ArcFace. 
    \item \textbf{SphereFace:} Liu \etal \cite{sphereface} proposed a deep hypersphere embedding approach (SphereFace) for FR task. SphereFace \cite{sphereface} utilized the angular softmax loss for CNNs to learn angularly discriminative features. This method also achieved competitive performance on LFW \cite{LFWTech} (accuracy of 99.42\%) and YTF \cite{YTF} datasets (95.00\%). 
    We \mf{extract the face representation with 512-dimension from a pre-trained 20-layer SphereFace model. \footnote{The official SphereFace model: \url{https://github.com/wy1iu/sphereface}}}
    \item \textbf{VGGFace2:} The first version of VGGFace is based on 16-layer VGG \cite{vgg16} network, while the second version of VGGFace (VGGFace2) \cite{vggface2} adopt ResNet-50 \cite{resnet} as the backbone architecture. 
    \mf{In this work}, we use the second version that a ResNet-50 network trained on VGGFace2 dataset \cite{vggface2} \footnote{The VGGFace2 model: \url{https://github.com/WeidiXie/Keras-VGGFace2-ResNet50}} for extracting the 512-dimension templates. 
\end{itemize}
The vulnerability of each FR system \pr{to} attacks \pr{was} analyzed based on three scenarios. Regardless of the scenario, the references are scenarios-specific bona fide videos captured \pr{on} the first day, while bona fide videos from the second and third days or attack videos \pr{were} selected as probes. The three cases, including the division of scenario-specific references and probes, are described with the results in \mf{detail} in Sec.~\ref{sec:fr_results}. Once the references for the face images are obtained, we use the Cosine-similarity as recommended in \cite{arcface, sphereface, vggface2} to compute the similarity scores between references and probes.

\section{Analysis of PAD Performance}
\label{sec:pad_results}
\mf{This section first describes three protocols designed to investigate the effect of masked attacks on PAD performance under different training settings. Second, the PAD evaluation metrics \pr{used were} introduced for further analysis. Third, quantitative results \pr{were} reported according to different PAD protocols. Finally, qualitative analysis and visualization are presented and discussed.}

\subsection{PAD evaluation protocols}
\label{ssec:pad_protocols}
\subsubsection{PAD protocols for the CRMA database}
\label{sssec:pad_protocols_crma}
\mf{In this \pr{study}, three protocols are provided to study the impact of masks on the performance of PAD solutions under different training settings. \pr{Other} factors, such as various devices, illumination, and capture scales, are \pr{outside the} scope \pr{of} this study. These three protocols try to answer three questions separately: 1) Does the PAD algorithm trained on unmasked data generalize well on the masked bona fides and attacks, that is, can the previously trained model be adapted to the present-day situation? 2) Does the PAD algorithm designed before the COVID-19 pandemic still \pr{work} efficiently if it is trained on additional masked data? 3) Will a network that has learned masked face attacks be confused by real masks that obscure the spoof face? }
\mf{Hence, we} split 47 subjects in the CRMA database into three subject-disjoint sets: the training set (19 subjects), the development set (10 subjects), and the testing set (18 subjects). Gender was balanced as much as possible between the three sets. Tab.~\ref{tab:protocols_description} provides more information about three protocols. \pr{A} detailed description of three protocols \pr{is as follows}: \\
\textbf{Protocol-1 (P1)}: \mf{This protocol demonstrates the generalization performance of the PAD solutions trained on unmasked data. The training and development sets contain only videos of subjects without masks (\pr{such as} data in most current PAD databases). The trained model \pr{was} then tested on the data \pr{using} face masks. More specifically, only BM0 and AM0 data \pr{were} used for training, while BM1, AM1, and AM2 were considered unknown mask data.} \\
\textbf{Protocol-2 (P2)}: \mf{In contrast to protocol-1, which focuses on generalizability on unseen mask data, the second protocol is designed to evaluate the performance of PAD algorithms when masked data has been learned in the training phase. In this protocol, the training, development, and testing sets include masked and unmasked bona fides (BM0, BM1), masked and unmasked attacks (AM0, AM1), and spoof faces with real masks (AM2).} \\
\textbf{Protocol-3 (P3)}: Until now, the effect of \mf{AM2} on PAD performance is still unclear. \mf{AM2 is a special attack type that a real face mask is placed on spoof faces, which means it contains only partial artifacts (i.e., unmasked face spoofing region) compared to AM1, which carries entire artifacts (i.e., spoofed face and mask). Therefore, this protocol attempts to answer the \pr{following} question: \pr{If} the network has learned the masked attacks AM1, can this trained model not be confused by a real mask and perform similarly on the attack covered by a real mask AM2? Consequently, the training and development sets include bona fides BM0 and BM1, and attacks AM0 and AM1, while AM2 is \pr{an} unknown attack in the testing set.} 
Because data in the CRMA \pr{are} video sequences and the number of videos between bona fide and attack classes are imbalanced, we sampled 60 frames from a bona fide video and \mf{five} frames from an attack video to reduce data bias. In addition to different frame sampling, we also adapt the class weights inversely proportional to the class frequencies to reduce overfitting in the training phase. In the test phase, a final classification decision was determined by averaging the prediction scores of all sampled frames.

\begin{table*}[htbp]
\centering
\footnotesize
\resizebox{0.8\textwidth}{!}{%
\begin{tabular}{|c|c|c|c|c|c|}
\hline
Protocol & Set & Subjects & Types of masks & \# BF videos & \# Attack videos \\ \hline
\multirow{3}{*}{P1} & Train & 1-19 & \mf{BM0, AM0} & 57 & 1569 \\ \cline{2-6} 
 & Dev & 20-29 & \mf{BM0, AM0} & 30 & 810 \\ \cline{2-6} 
 & Test & 30-47 & \mf{BM0, BM1, AM0, AM1, AM2} & 162 & 4860 \\ \hline \hline
\multirow{3}{*}{P2} & Train & 1-19 & \mf{BM0, BM1, AM0, AM1, AM2} & 171 & 5130  \\ \cline{2-6} 
 & Dev & 20-29 & \mf{BM0, BM1, AM0, AM1, AM2} & 90 & 270  \\ \cline{2-6} 
 & Test & 30-47 & \mf{BM0, BM1, AM0, AM1, AM2}  & 162 & 4860\\ \hline \hline
\multirow{3}{*}{P3} & Train & 1-19 & \mf{BM0, BM1, AM0, AM1} & 171 & 4617 \\ \cline{2-6} 
 & Dev & 20-29 & \mf{BM0, BM1, AM0, AM1}  & 90 & 2430 \\ \cline{2-6} 
 & Test & 30-47 & \mf{BM0, BM1, AM0, AM1, AM2} & 162 & 4860  \\ \hline
\end{tabular}}
\caption{The detailed information of three protocols for exploration of the possible effect of face masks. The bona fide is denoted as BF. \mf{The test data is the same in the three protocols, while the types of training and development data are different.}}
\label{tab:protocols_description}
\end{table*}

\subsubsection{PAD protocols for cross-database scenarios}
\label{ssec:cross_db_settings}
In addition to the intra-database scenario on our CRMA database, we also perform cross-database experiments to explore the generalizability of these PAD algorithms on masked data. Because the PAIs in the CRMA database are print and replay attacks, we \pr{selected} three popular publicly available databases containing the same PAIs: CASIA-MFS \cite{casia_fas}, MSU-MFS \cite{msu_mfs}, and OULU-NPU \cite{oulu_npu} to demonstrate the evaluation. We \pr{conducted} two cross-database experiments. In the first cross-database scenario, the PAD solutions trained on three publicly available databases were evaluated on the test set of the CRMA database. In addition, the results tested on their own test sets are also reported (as shown in the left block in Tab.~\ref{tab:cross_db_1}). The first setting is similar to protocol-1 of the CRMA intra-database scenario, as no masked data \pr{are} seen in the training phase. Therefore, the first cross-database setting is also used to answer the first question: does the PAD algorithm trained on unmasked data generalize well on masked bona fides and attacks? 
Conversely, in the second cross-database experiment, models trained on different protocols of the CRMA database were evaluated separately on publicly available databases. This experimental setting can help us understand the CRMA database values beyond face masks, such as the diversity of masks/sensors/scales. However, the second scenario does not support the main study of the work and is provided only for completeness; thus, the results are reported in \pr{the} supplementary material.  
In both cross-database scenarios, we use the $\tau_{BPCER10}$ \pr{decision threshold} computed on the development set of the training database as \pr{a} priori to determine the APCER, BPCER, and HTER \pr{values} of the test database.

\subsection{PAD evaluation metrics}
The metrics following the ISO/IEC 30107-3 \cite{ISO301073} standard \pr{were} used to measure the performance of the PAD algorithms: \textit{\mf{Attack} Presentation Classification Error Rate} (APCER) and \textit{bona fide presentation classification error rate } (BPCER). APCER is the proportion of attack images incorrectly classified as bona fide samples in a specific scenario, while BPCER is the proportion of bona fide images incorrectly classified as \mf{attacks} in a specific scenario. The APCER and BPCER reported in the test set were based on a pre-computed threshold in the development set. In our study, we use a BPCER at 10\% (on the development set) \pr{to obtain} the threshold (denoted as $\tau_{BPCER10}$). Additionally, \textit{ half-total error rater} (HTER) corresponding to half of the summation of BPCER and APCER is used for the cross-database evaluation. Noticeably, we \pr{computed} a threshold in the development set of the training database. Then, this threshold was used to determine the HTER value in the test database. The detection EER (D-EER) value, where APCER and BPCER are equal \mf{is} also reported in the cross-database scenarios. For further analysis of PAD performance, receiver operating characteristic (ROC) curves \pr{were} also demonstrated.

\begin{table*}[htb!]
\footnotesize
\centering
\def\arraystretch{1.0}
\resizebox{0.9\textwidth}{!}{%
\begin{tabular}{|c|c|c|c||c|c|c||c|c|c|}
\hline
\multirow{3}{*}{Protocol} & \multirow{3}{*}{Method} & \multicolumn{8}{c|}{Threshold @ BPCER 10\% in dev set} \\ \cline{3-10} 
 &  & \multicolumn{2}{c|}{BPCER (\%)} & \multicolumn{3}{c|}{APCER (print) (\%)} & \multicolumn{3}{c|}{APCER (replay) (\%)} \\ \cline{3-10} 
 &  & \mf{BM0} & \mf{BM1} & \mf{AM0} & \mf{AM1} & \mf{AM2} & \mf{AM0} & \mf{AM1} & \mf{AM2} \\ \hline
\multirow{5}{*}{P1} & LBP & 1.75 & \textbf{4.39} & \textbf{80.12} & 72.61 & 71.93 & \textbf{74.95} & 67.76 & 73.98 \\ \cline{2-10} 
 & $\mathrm{Inception_{FT}}$ & 19.30 & \textbf{84.21} & \textbf{10.33} & 3.80 & 2.92 & \textbf{27.19} & 5.81 & 0.88 \\ \cline{2-10} 
 & CPqD & 7.02 & \textbf{47.37} & \textbf{18.52} & 7.80 & 15.79 & \textbf{31.77} & 11.19 & 10.23 \\ \cline{2-10} 
 & $\mathrm{FASNet_{FT}}$ & 12.28 & \textbf{56.14} & \textbf{7.02} & 1.36 & 2.92 & \textbf{20.37} & 12.21 & 9.65 \\ \cline{2-10} 
  & $\mathrm{Inception_{TFS}}$ & 7.04 & \textbf{48.25} & 1.36 & 0.00 & \textbf{1.75} & \textbf{7.50} & 0.34 & 7.02 \\ \cline{2-10}
 & $\mathrm{FASNet_{TFS}}$ & 7.02 & \textbf{29.82} & 1.95 & 0.49 & \textbf{15.20} & \textbf{8.09} & 4.64 & 7.89 \\ \cline{2-10}
 & DeepPixBis & 19.30 & \textbf{28.95} & 1.56 & 1.56 & \textbf{5.85} & 3.61 & 4.05 & \textbf{6.43} \\ \hline \hline
\multirow{5}{*}{P2} & LBP & \textbf{26.32} & 11.40 & 31.38 & \textbf{44.44} & 36.84 & \textbf{36.74} & 34.39 & 28.95 \\ \cline{2-10} 
 & $\mathrm{Inception_{FT}}$ & 1.75 & \textbf{7.02} & \textbf{35.28} & 30.80 & 11.70 & \textbf{54.09} & 52.17 & 10.23 \\ \cline{2-10} 
 & CPqD & 3.51 & \textbf{7.89} & 27.49 & \textbf{30.41} & 16.37 & \textbf{46.20} & 44.50 & 10.23 \\ \cline{2-10} 
 & $\mathrm{FASNet_{FT}}$ & 1.75 & \textbf{17.54} & 10.72 & \textbf{12.77} & 5.85 & \textbf{30.60} & 28.09 & 3.80 \\ \cline{2-10}
 & $\mathrm{Inception_{TFS}}$ & 8.77 & \textbf{18.42} & 0.78 & 1.56 & \textbf{2.34} & 3.90 & \textbf{5.23} & 2.63 \\ \cline{2-10}
 & $\mathrm{FASNet_{TFS}}$ & 14.04 & \textbf{29.82} & 4.09 & 3.41 & \textbf{9.36} & \textbf{4.69} & 2.88 & 3.80 \\ \cline{2-10}
 & DeepPixBis & \textbf{29.82} & 24.56 & 0.78 & 0.19 & \textbf{1.75} & 0.10 & \textbf{1.86} & 0.88 \\ \hline \hline
\multirow{5}{*}{P3} & LBP & \textbf{22.81} & 9.65 & 35.28 & \textbf{48.15} & 47.95 & 38.50 & 36.79 & \textbf{42.40} \\ \cline{2-10} 
 & $\mathrm{Inception_{FT}}$ & 1.75 & \textbf{8.77} & 24.17 & \textbf{24.37} & 11.70 & 46.69 & \textbf{47.14} & 14.04 \\ \cline{2-10} 
 & CPqD & \textbf{7.02} & \textbf{7.02} & 20.66 & \textbf{28.95} & 21.64 & 41.23 & \textbf{41.52} & 17.84 \\ \cline{2-10} 
 & $\mathrm{FASNet_{FT}}$ & 5.26 & \textbf{21.93} & 14.04 & 9.94 & \textbf{26.71} & \textbf{22.62} & 19.88 & 20.47 \\ \cline{2-10} 
 & $\mathrm{Inception_{TFS}}$ & 21.05 & \textbf{21.93} & 0.19 & 0.00 & \textbf{1.17} & 1.56 & 2.34 & \textbf{4.97} \\ \cline{2-10}
 & $\mathrm{FASNet_{TFS}}$ & 22.81 & \textbf{34.21} & 0.39 & 0.29 & \textbf{2.34} & 3.41 & 2.20 & \textbf{6.43} \\ \cline{2-10}
 & DeepPixBis & 17.54 & \textbf{24.56} & 0.78 & 0.68 & \textbf{2.92} & 0.88 & 1.91 & \textbf{6.43} \\ \hline
\end{tabular}}
\caption{\mf{The PAD performance of different PAD solutions in three protocols (as described in Sec.\ref{ssec:pad_protocols}). The bold number in each protocol and each method refers to the highest BPCER on BM0 and BM1 data and \pr{the} highest APCER value between AM0, AM1, and AM2 in \pr{the} two PAIs, respectively. The higher BPCER values for BM1 (in comparison to BM0) indicate that subjects wearing masks tend to be classified falsely as attacks.}}
\label{tab:intra_db_test}
\end{table*}

\begin{figure*}[htbp!]
\centering
\subfloat[ROC curves in protocol-1]{
	\label{subfig:p1_roc}
	\includegraphics[width=0.9\textwidth]{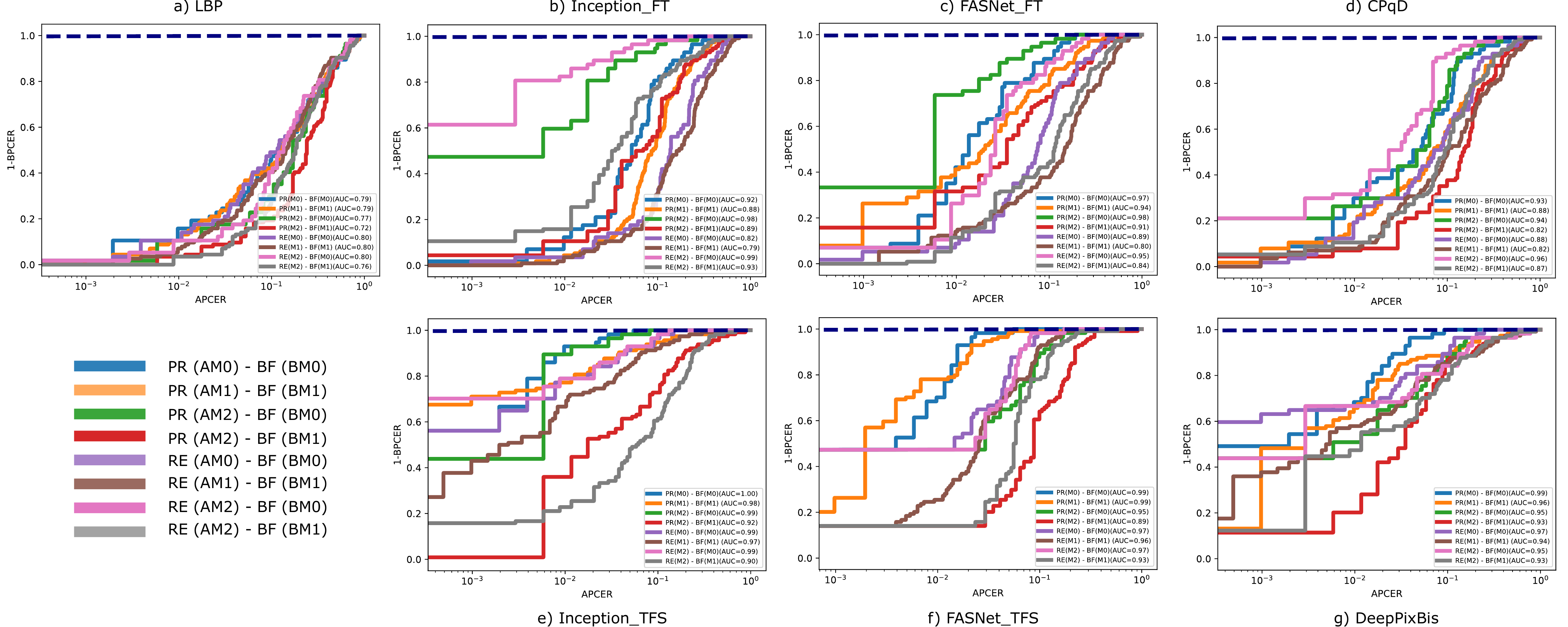}} 

\subfloat[ROC curves in protocol-2]{
	\label{subfig:p2_roc}
	\includegraphics[width=0.9\textwidth]{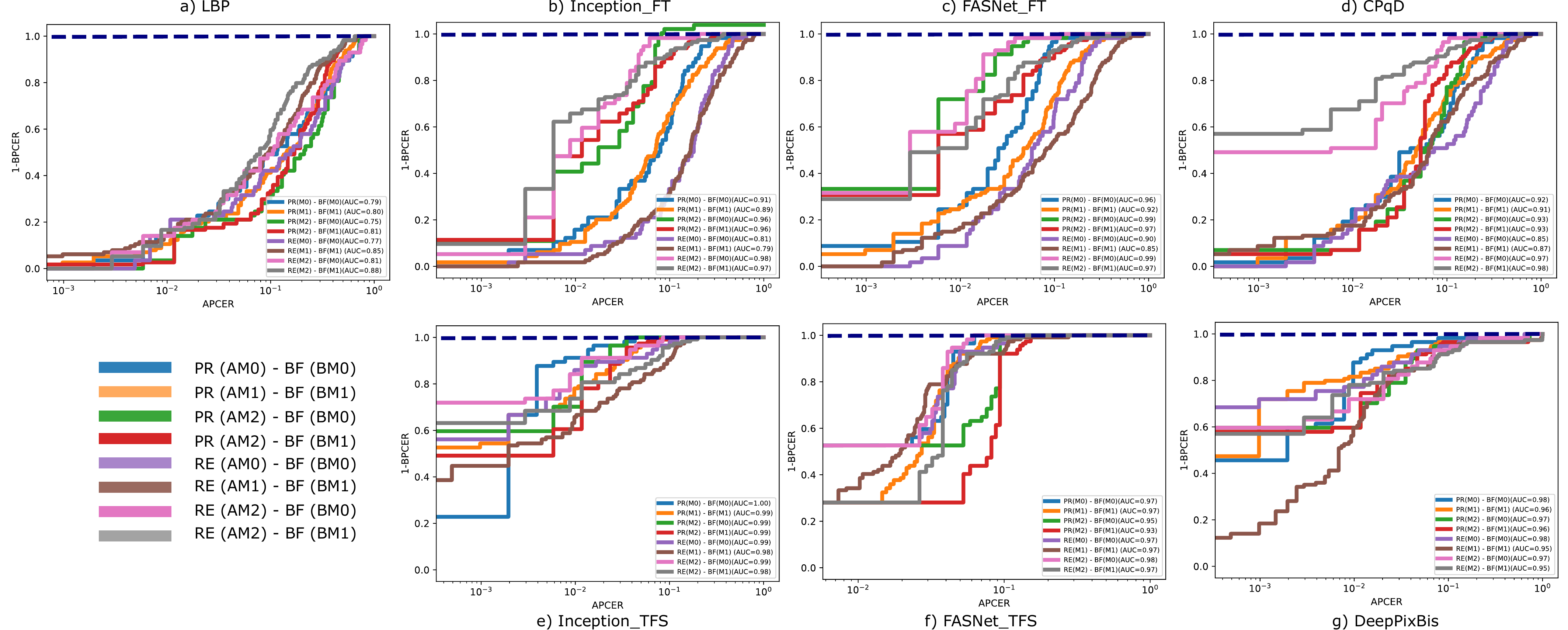} } 
 
\subfloat[ROC curves in protocol-3]{
	\label{subfig:p3_roc}
	\includegraphics[width=0.9\textwidth]{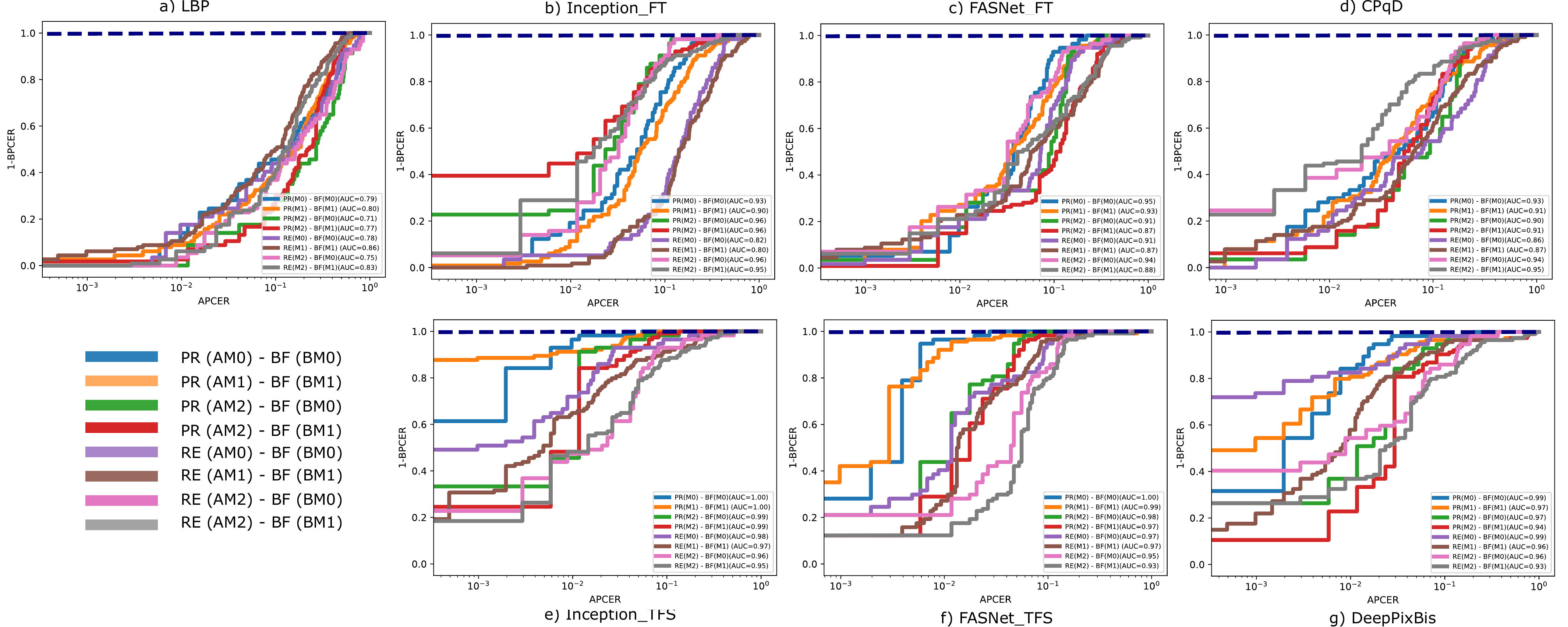}} 
\vspace{-3mm}
\caption{ROC curves for all PAD methods in three protocols. \mf{Eight combinations between bona fide and attack (testing data) are represented for each method in each protocol: PR(AM0)-BF(BM0), PR(AM1)-BF(BM1), PR(AM2)-BF(BM0), PR(AM2)-BF(BM1) in print PAI and RE(AM0)-BF(BM0), RE(AM1)-BF(BM1), RE(AM2)-BF(BM0), RE(AM2)-BF(BM1) in replay PAI. The x-axis and y-axis are APCER and 1 - BPCER, respectively.} The red curves (PR(AM2)-BF(BM1)) and gray curves (RE(AM2)-BF(BM1)) \pr{show} significantly smaller AUC values by most PAD methods on protocol-1. Moreover, $\mathrm{Inception_{TFS}}$, $\mathrm{FASNet_{TFS}}$, and DeepPixBis achieve higher AUC values on \mf{protocol-2 and -3 than on protocol-1 might be due to \pr{the} masked data in the training phase.}}
\label{fig:intra_db_roc}
\end{figure*}

\subsection{Analysis of Protocol-1}
\mf{Protocol-1} represents the pre-COVID-19 PAD scenarios, in which subjects normally do not wear a mask, \mf{and demonstrates the generalization performance on masked data}. Therefore, protocol-1 is considered the most challenging task \pr{because of} the unseen BM1, AM1, and AM2 data. 
\mf{Tab.~\ref{tab:intra_db_test} describes the results of \pr{the} different protocols \pr{on} the CRMA database. The bold numbers indicate the highest BPCER values between BM0 and BM1 and the highest APCER values between AM0, AM1, and AM2 in each PAI. By observing the first block, P1, in Tab.~\ref{tab:intra_db_test},} the BPCER values of masked bona fide samples are much higher than \pr{those of} unmasked ones; \pr{however}, most PAD systems achieve \mf{lower} APCER values on the masked attack samples (either \mf{AM1 or AM2}). \mf{The higher classification error rates on masked bona fide and the lower error rates on masked attacks are intuitively conceivable. When the model has not seen faces wearing a mask before, it is more inclined to falsely classify such a masked bona fide sample (BM1) as an attack.}
Moreover, it is interesting to note that networks trained from scratch and the DeepPixBis approach work worse on attack \mf{AM2 than AM1}. These observations are consistent with the ROC (Fig.~\ref{fig:intra_db_roc}). The red curves generated by printed \mf{AM2}, bona fide \mf{BM1}, and gray curves obtained by replay \mf{AM2} and bona fide \mf{BM1} possess significantly smaller areas under the curves in five of the seven methods. Furthermore, training a network from \mf{scratch} \pr{improves} the overall performance. \mf{The possible reason for those observations is} that learning from scratch is more efficient for obtaining discriminative features between bona fide and artifacts. On the contrary, such approaches might be confusing when applying realistic masks \pr{to} attack samples.

\mf{In addition to the intra-database scenario, the first cross-database experiment (introduced in Sec.~\ref{ssec:cross_db_settings}) can be seen \pr{as} similar to protocol-1, as both scenarios study PAD methods that PAD solutions trained on unmasked data and tested on the CRMA database. In Tab.~\ref{tab:cross_db_1}, the bold BPCER number is the highest BPCER (between BM0 and BM1) for each PAD method. The bold APCER number is the highest APCER (between AM0, AM1, and BM2) for each PAD method in print and replay attacks, respectively. This bolding is performed to show which samples are \pr{more difficult} to \pr{classify} correctly. We observed that the performance in the cross-database setting \pr{was} relatively poor for all models. Even though deep-learning-based methods achieved great results on their own test sets, they generalize significantly worse on masked bona fide samples; \pr{for example}, most BPCER values for BM1 are close to 100\%. \pr{In contrast}, most algorithms achieve lower APCER values on masked AM1 and AM2 than unmasked AM0 attacks, \pr{which is} consistent with the observation of protocol-1 from the intra-database scenarios.} 

\begin{table*}[htb!]
\footnotesize
\centering
\def\arraystretch{1.1}
\resizebox{\textwidth}{!}{
\begin{tabular}{|c|c|c|c|c||c|c||c|c|c||c|c|c|}
\hline
\multirow{2}{*}{Trained on} & \multirow{4}{*}{Method} & \multicolumn{11}{c|}{Threshold @ BPCER 10\% in dev set of trained database} \\ \cline{3-13} 
 &  & \multicolumn{3}{c|}{Tested on the same dataset (\%) } & \multicolumn{8}{c|}{Tested on our CRMA dataset (\%)} \\ \cline{1-1} \cline{3-13} 
\multirow{7}{*}{CAISA-FASD} &  & \multirow{2}{*}{D-EER} & \multirow{2}{*}{BPCER} & \multirow{2}{*}{APCER } & \multicolumn{2}{c|}{BPCER} & \multicolumn{3}{c|}{APCER (Print)} & \multicolumn{3}{c|}{APCER (Replay)} \\ \cline{6-13} 
 &  &  &  &  & \mf{BM0} & \mf{BM1} & \mf{AM0} & \mf{AM1} & \mf{AM2} & \mf{AM0} & \mf{AM1} & \mf{AM2} \\ \cline{2-13} 
 & LBP & 7.50 & 6.25 & 8.75 & 38.60 & \textbf{56.14} & \textbf{42.11} & 24.76 & 18.13 & \textbf{60.72} & 34.59 & 22.51 \\ \cline{2-13} 
 & $\mathrm{Inception_{FT}}$ & 10.00 & 8.75 & 15.00 & 21.05 & \textbf{38.60} & \textbf{35.48} & 5.95 & 16.96 & \textbf{69.49} & 47.44 & 15.50 \\ \cline{2-13} 
 & CPqD & 6.25 & 11.25 & 3.12 & 38.60 & \textbf{65.79} & \textbf{31.97} & 12.38 & 8.77 & \textbf{53.22} & 23.06 & 14.62 \\ \cline{2-13} 
 & $\mathrm{FASNet_{FT}}$ & 8.75 & 12.50 & 4.38 & 15.79 & \textbf{90.35} & \textbf{44.83} & 2.14 & 23.98 & \textbf{64.13} & 5.76 & 22.81 \\ \cline{2-13} 
 & $\mathrm{Inception_{TFS}}$ & \textit{0.00} & \textit{1.25} & \textit{0.00} & 12.28 & \textbf{20.08} & \textbf{61.60} & 40.35 & 49.71 & \textbf{90.35} & 83.19 & 59.65 \\ \cline{2-13} 
  & $\mathrm{FASNet_{TFS}}$ & 1.25 & 3.75 & 0.62 & 21.05 & \textbf{75.44} & \textbf{60.23} & 19.49 & 38.60 & \textbf{70.86} & 16.32 & 45.61 \\ \cline{2-13}
 & DeepPixBis & 1.25 & 6.25 & \textit{0.00} & 35.09 & \textbf{66.67} & \textbf{70.57} & 36.65 & 56.73 & \textbf{57.99} & 29.26 & 42.98 \\ \hline \hline
\multirow{5}{*}{MSU-MFSD} & LBP & 4.17 & \textit{4.17} & 4.17 & 98.25 & \textbf{100.00} & 0.58 & \textbf{0.68} & 0.00 & \textbf{3.22} & 2.25 & 0.00 \\ \cline{2-13} 
 & $\mathrm{Inception_{FT}}$ & 20.14 & 20.81 & 16.67 & \textbf{50.88} & 25.44 & 47.95 & \textbf{56.04} & 52.05 & 31.19 & \textbf{48.85} & 44.15 \\ \cline{2-13} 
 & CPqD & 4.17 & \textit{4.17} & 4.17 & 98.25 & \textbf{100.00} & 0.19 & \textbf{0.39} & 0.00 & \textbf{1.46} & 1.56 & 0.00 \\ \cline{2-13} 
 & $\mathrm{FASNet_{FT}}$ & 13.19 & 26.39 & 4.17 & 43.86 & \textbf{85.96} & \textbf{32.55} & 2.63 & 0.58 & \textbf{42.50} & 13.39 & 2.34 \\ \cline{2-13}
  & $\mathrm{Inception_{TFS}}$ & 4.17 & 8.33 & 1.39 & 80.70 & \textbf{94.74} & \textbf{0.19} & 0.00 & 0.00 & \textbf{8.58} & 0.78	& 2.05 \\ \cline{2-13} 
  & $\mathrm{FASNet_{TFS}}$ & \textit{0.00} & 8.44 & \textit{0.00} & 91.23 & \textbf{100.00}
  & 0.00 & 0.00 & 0.00 & \textbf{7.70} & 0.00 & 0.29 \\ \cline{2-13}
 & DeepPixBis & \textit{0.00} & \textit{4.17} & \textit{0.00} & \textbf{82.46} & 80.70 & 0.00 & \textbf{0.10} & 0.00 & 10.33 & \textbf{10.36} & 5.26 \\ \hline \hline
\multirow{5}{*}{Oulu-NPU} & LBP & 8.33 & 7.50 & 10.21 & 40.35 & \textbf{67.54} & \textbf{35.28} & 25.54 & 13.45 & \textbf{26.12} & 10.89 & 13.74 \\ \cline{2-13} 
 & $\mathrm{Inception_{FT}}$ & 15.00 & 16.67 & 11.04 & 61.40 & \textbf{87.72} & \textbf{11.50} & 5.85 & 8.77 & \textbf{12.38} & 2.39 & 1.46 \\ \cline{2-13} 
 & CPqD & 8.33 & 9.17 & 3.54 & 57.89 & \textbf{89.47} & \textbf{9.55} & 3.70 & 1.17 & \textbf{10.14} & 1.03 & 0.58 \\ \cline{2-13} 
 & $\mathrm{FASNet_{FT}}$ & 3.23 & \textit{1.67} & 4.38 & 49.12 & \textbf{73.68} & \textbf{33.92} & 27.10 & 8.77 & \textbf{22.81} & 8.99 & 3.80 \\ \cline{2-13} 
  & $\mathrm{Inception_{TFS}}$ & 4.17 & 3.33 & 6.46 & 80.07 & \textbf{100.00} & \textbf{22.81} & 0.78 & 2.34 & \textbf{3.22} & 0.00 & 0.00  \\ \cline{2-13} 
  & $\mathrm{FASNet_{TFS}}$ & 5.10 & 11.67 & 3.33 & 70.18 & \textbf{99.12} & \textbf{46.98} & 18.03 & 19.88 & \textbf{8.09} & 0.39 & 0.29 \\ \cline{2-13}
 & DeepPixBis & \textit{2.29} & 2.92 & \textit{0.00} & 66.67 & \textbf{98.25} & \textbf{44.64} & 11.21 & 4.68 & \textbf{10.23} & 0.10 & 0.58 \\ \hline
\end{tabular}}
\caption{Cross-database evaluation 1: the model trained on three publicly available databases is used to test on the CRMA database. \mf{This cross-database scenario is similar to protocol-1, as no masked data is seen during the training phase. Italic numbers indicate the lowest error rate on their own test set, and bold numbers indicate the highest error rate in the bona fide and each PAI category. The results show that despite good performance on their own test set, these trained models do not generalize well \pr{to} masked bona fides and attacks.}}
\label{tab:cross_db_1}
\end{table*}

\mf{In general, the experimental results of the intra-database protocol-1 and the first cross-database scenario results answer the first posed question (in Sec.~\ref{sssec:pad_protocols_crma}) by showing that models trained only on unmasked data cannot properly classify images of masked faces. A subject with a mask on has a high probability of being falsely detected as an attack by PAD systems, even if this subject is bona fide. }

\subsection{Analysis of Protocol-2}
\mf{Protocol-2 targets the performance of PAD algorithms on masked data when both unmasked and masked samples are used in the training phase.}
As shown in Tab.~\ref{tab:intra_db_test}, we can observe the following points: First, despite \pr{the fact} that the masked bona fide samples are still more difficult to classify correctly than unmasked ones in most cases, \mf{the BPCER value of BM1 behaves more similar to its behavior on the BM0 in protocol-2 than in protocol-1. Moreover, the BPCER values of BM0 and BM1 in protocol-2 \pr{decreased} in most cases compared \pr{with} the results of protocol-1. For example, the BPCER value of BM1 achieved by $\mathrm{Inception_{FT}}$ \pr{was} 84.21\% in protocol-1 and 7.02\% in protocol-2. This finding indicates that learning \pr{the} masked data is helpful \pr{in improving} the performance of PAD methods.} 
This is also \pr{consistent} with the observation in the ROC curves \mf{(by comparing the ROCs in protocol-1 and protocol-2 in general)}. In particular, $\mathrm{Inception_{TFS}}$, $\mathrm{FASNet_{TFS}}$, and DeepPixBis \pr{achieved} significant progress (larger areas under the curves). Second, six of the seven methods \pr{performed} worse on the masked printed face \mf{(AM1 or AM2)}, while five of the seven algorithms showed inferior results for unmasked replay attacks. Moreover, \mf{AM2} in print PAI achieves higher APCER values than \mf{AM1} by training from scratch approaches. One possible reason for the different results between print and replay attacks is specular reflection. Because attack data were collected in windowless labor with all electric lights on, tablets easily reflect the light \pr{compared to} the printed paper, and this reflection is difficult to avoid. The \mf{real} face masks \pr{might also} leak light when placed on an electric tablet, but this does not appear when applied on printed paper. 
\mf{In general, the experimental results of the intra-database protocol-2 answer the second question (in Sec.~\ref{sssec:pad_protocols_crma}), which \pr{addresses} the performance changes of \pr{the} current PAD algorithms after complementary learning on \pr{the} masked data. \pr{Based on} the above findings, we can conclude that the PAD algorithms still perform worse on masked bona fides (BM1) than on unmasked faces (BM0 ), even when the PAD solutions are trained on masked data.}

\subsection{Analysis of Protocol-3}

\mf{Protocol-3 investigates the generalizability of the model trained on data that includes masked face attacks (AM1) when tested on the masked face attacks where a real mask is placed on top of the attack (AM2).}
For bona fide samples, we draw a similar conclusion to \mf{protocol-1 and protocol-2, stating that masked bona fide samples have a higher probability of incorrectly being classified as attacks. However, the experimental results show differences in attack detection \pr{behavior} (APCER) between protocol-3 on one side and protocols-1 and -2 on the other side.} In this protocol, the highest APCER values of most PAD algorithms appear on either the \mf{AM1 or AM2} attacks in both print and replay PAIs. Second, the traditional LBP method, $\mathrm{Inception_{FT}}$, $\mathrm{FASNet_{FT}}$, and \pr{the} hybrid CPqD method that achieve relatively \mf{worse} results on \mf{AM0 or AM1 attacks than other methods} may have proved to be unable to learn or extract sufficient discriminative features.
\mf{Third,} although the \mf{other methods, such as} learning from \mf{scratch} $\mathrm{Inception_{TFS}}$ and $\mathrm{FASNet_{TFS}}$ or \mf{custom designed DeepPixBis} achieve impressive results on \mf{seen AM0 and AM1} attacks, they generalize not well on \mf{unseen AM2} attacks. \mf{These observations answer the third question stated in Sec.~\ref{sssec:pad_protocols_crma} by stating that a network trained on masked face attacks (AM1) tends to produce confusing decisions on AM2, where a real mask is placed on an attack face.} 
\subsection{Qualitative Analysis and Visualization}

\begin{figure*}[htbp!]
\centering
\subfloat[Protocol-1]{
	\label{subfig:p1_scorecam}
	\includegraphics[width=0.33\textwidth]{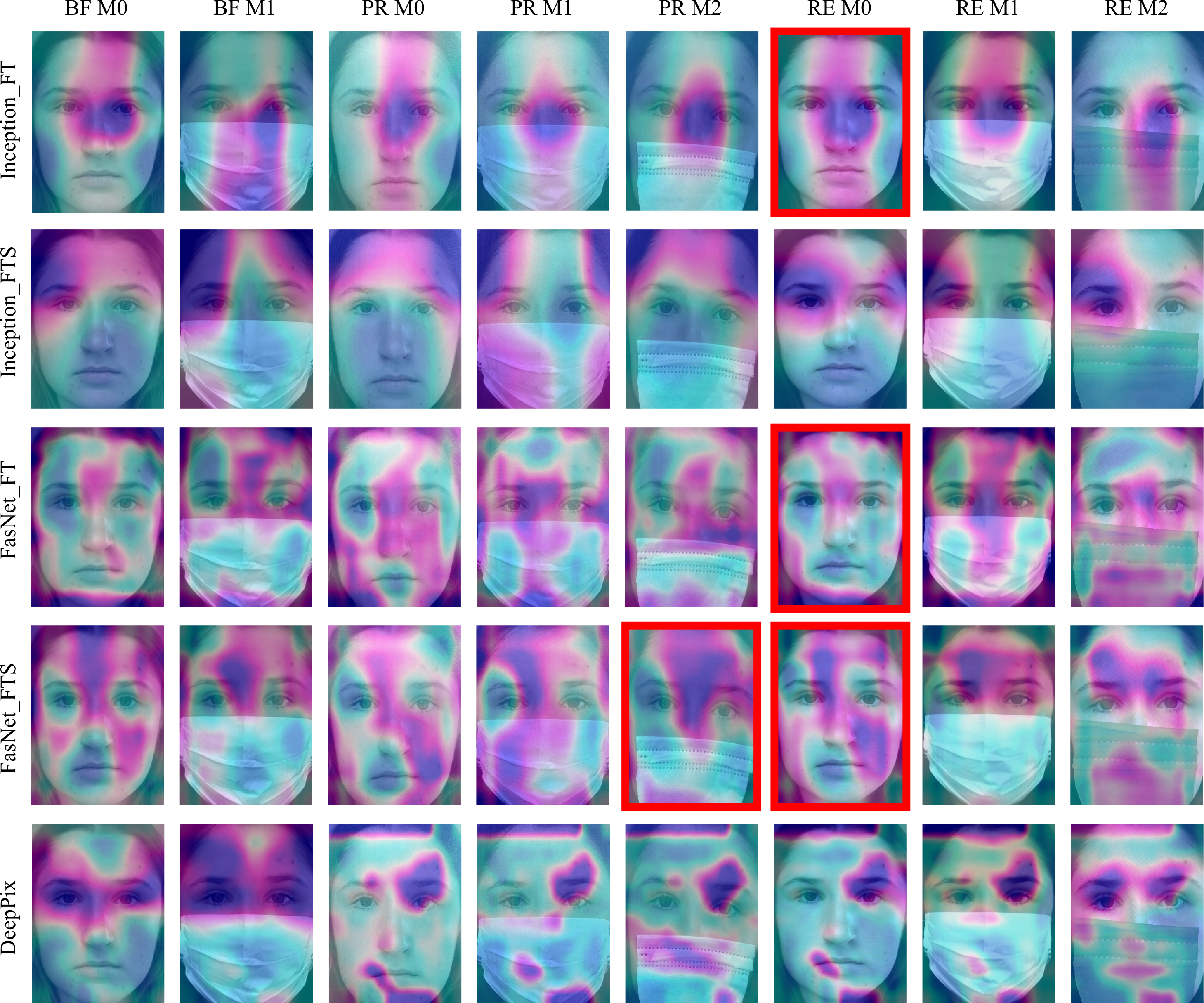}} 
\subfloat[Protocol-2]{
	\label{subfig:p2_scorecam}
	\includegraphics[width=0.33\textwidth]{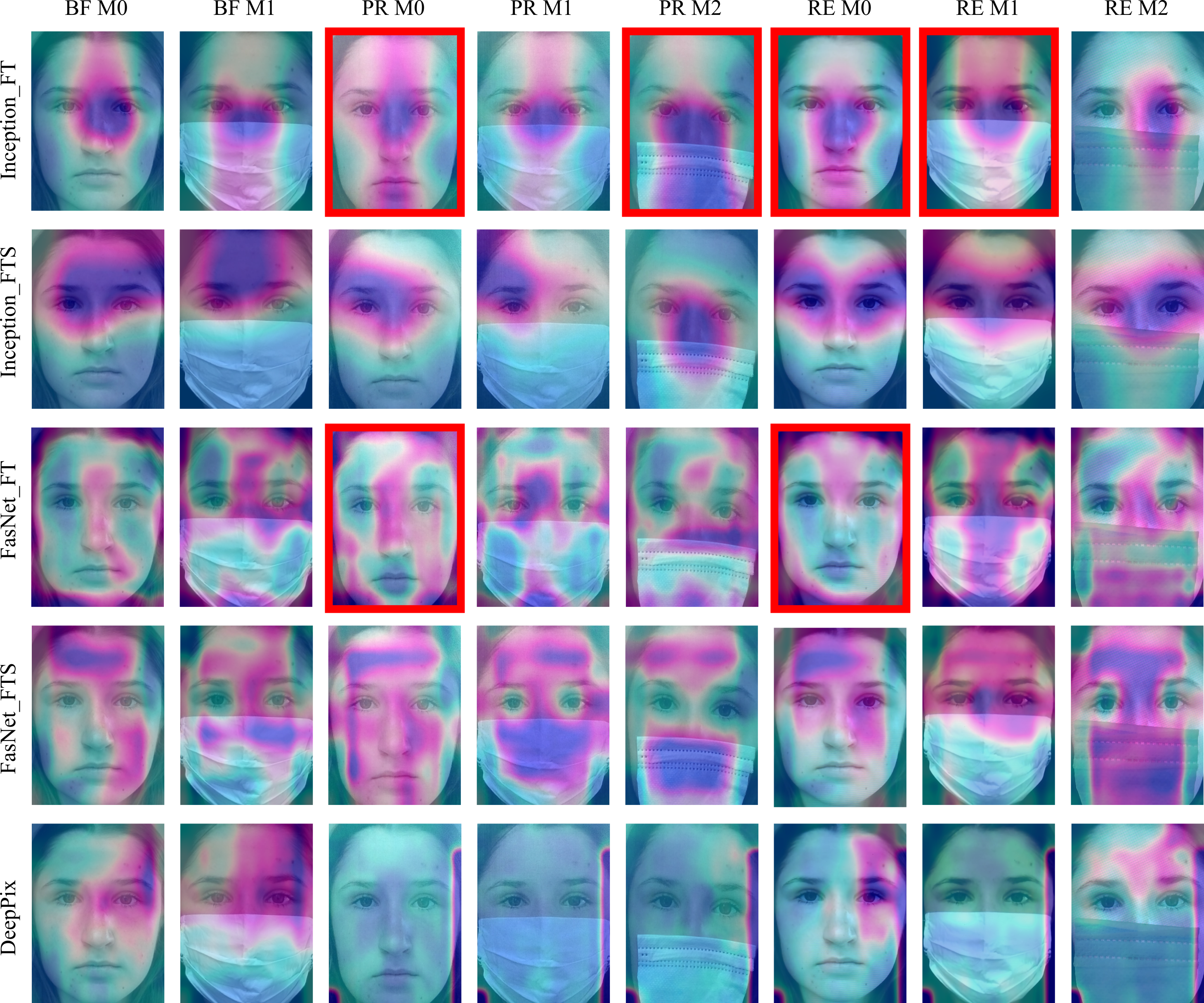}} 
\subfloat[Protocol-3]{
	\label{subfig:p3_scorecam}
	\includegraphics[width=0.33\textwidth]{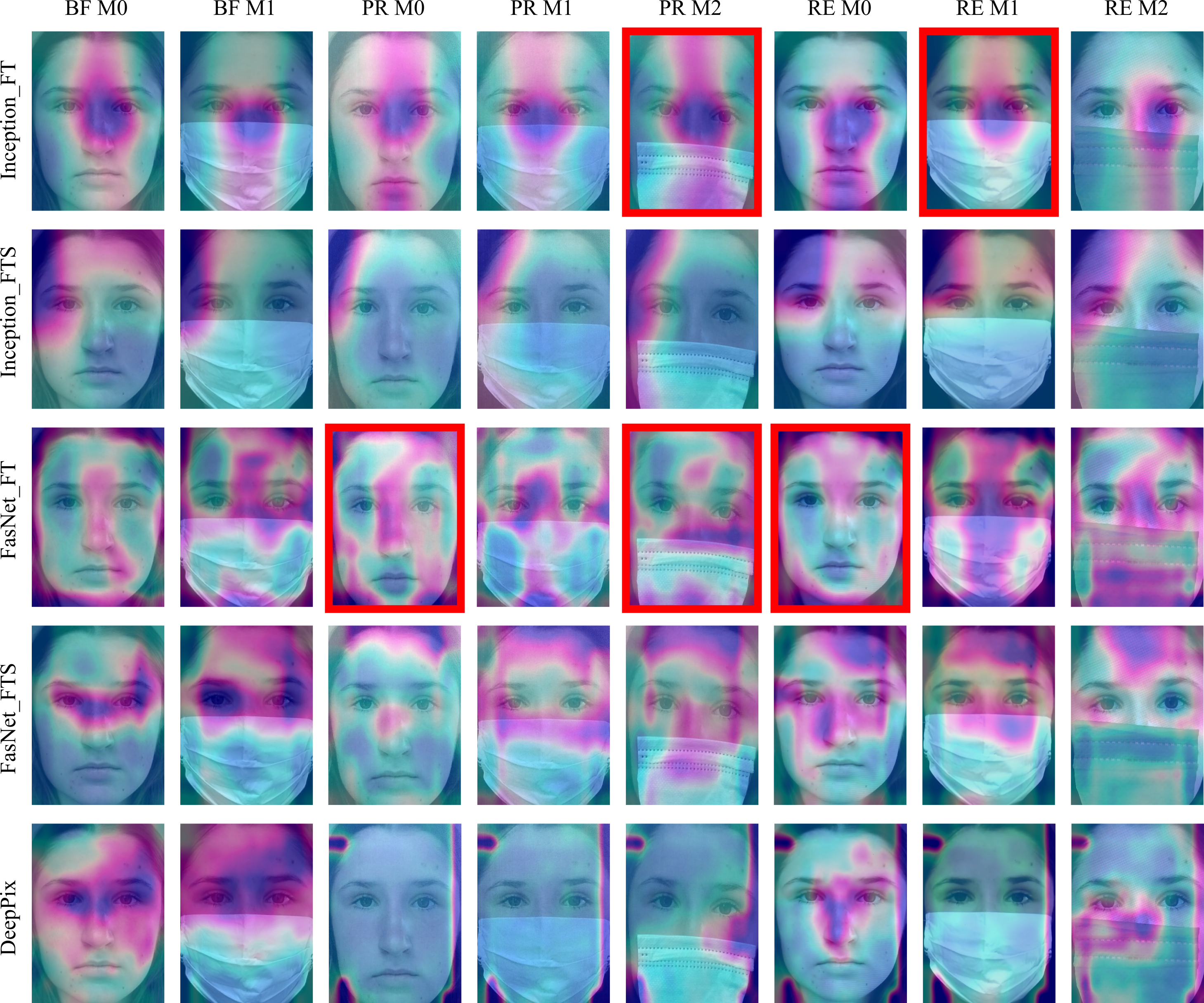}}
\caption{Examples for attention maps generated by ScoreCAM of different PAD algorithms and different protocols. The rows from top to bottom in each protocol correspond to $\mathrm{Inception_{FT}}$, $\mathrm{Inception_{TFS}}$, $\mathrm{FASNet_{FT}}$, $\mathrm{FASNet_{TFS}}$, and DeepPixBis. The columns from left to right in each \mf{protocol} refer to \mf{BM0, BM1, PR-AM0, PR-AM1, PR-AM2, RE-AM0, RE-AM1, RE-AM2.} Faces with red boxes are misclassified.}
\label{fig:scorecam}
\end{figure*}

To qualitatively analyze and interpret the deep-learning-based methods, the score-Weighted CAM \cite{DBLP:conf/cvpr/WangWDYZDMH20} technique \pr{was} adopted to localize the discriminative areas in face images. 
The rows from top to bottom correspond to $\mathrm{Inception_{FT}}$, $\mathrm{Inception_{TFS}}$, $\mathrm{FASNet_{FT}}$, $\mathrm{FASNet_{TFS}}$ and DeepPixBis. Fig.~\ref{subfig:p1_scorecam} shows the results of \mf{protocol-1} (the example subject is in the test set). $\mathrm{Inception_{FT}}$ mainly focuses on the nose, including nearby parts of the masks, \pr{whereas} $\mathrm{Inception_{TFS}}$ pays more attention to the upper region of the face. Similarly, $\mathrm{FASNet_{TFS}}$ reduces the attention \pr{paid to the} masks and increases the concentration around the forehead. DeepPixBis concentrates around the eyes for both \mf{unmasked (BM0) and masked (BM1) bona fides}. However, for attack samples, attention seems \pr{to be} focused on the left eye and partial masks. In general, masks are noticed by all networks. The results of \mf{protocol-2 and protocol-3} for the same subjects \pr{are shown} in Fig.~\ref{subfig:p2_scorecam}, and Fig.~\ref{subfig:p3_scorecam}. We noticed that 1) the attention areas of fine-tuned networks hardly change in \pr{the} three protocols \pr{because of} the fixed weights of layers before the last classification layer. 2) $\mathrm{Inception_{TFS}}$ in \mf{protocol-2} \pr{appears to focus} on the upper face, including \pr{many} more eye regions than in \mf{protocol-1}. 3) $\mathrm{FASNet_{TFS}}$ in \mf{protocol-2} concentrates much more on applied real masks than in \mf{protocol-3} where training without \mf{AM2}. 4) DeepPixBis still works well on bona fide, but for attack samples, its attention seems to be distracted to the edge of images. \pr{Although} DeepPixBis produces correct decisions, this observation raises a serious concern about its reliability and generalizability. This concern \pr{was} confirmed \pr{by} the cross-database evaluation. DeepPixBis generally obtains worse cross-database results than \pr{the} other two training from scratch networks (details see Tab.~\ref{tab:cross_db_1}). Finally, looking at attention maps in all protocols for this identity, we \mf{notice} that except for the misclassified samples (with red boxes) that appear on print/replay \mf{AM0}, print \mf{AM2} attacks are \pr{more easily} to be incorrectly detected as bona fide than \mf{AM1} attacks. \mf{This finding is in line with the previous quantitative evaluation that AM2 attacks may confuse the PAD, even if the network has been trained by masked face attacks.}  

To further understand the above quantitative and qualitative results, we provide additional t-SNE plots to visualize the learned features in the supplementary material. These plots consolidate our findings here that 1) masked bona fide samples are more \pr{likely} to be detected as bona fide by the pre-COVID-19 PAD algorithms. 2) attacks with real masks placed on the attacks (AM2) are more falsely detected by PAD systems as bona fides than attacks with masked faces (AM1).

\vspace{-3mm}
\section{Analysis of FR Vulnerability}
\label{sec:fr_results}
\subsection{Experimental settings} 
The vulnerability of each FR system on \mf{each type of PA} is analyzed based on three experimental settings. In the first setting \mf{BM0-BM0}, we use the bona fide unmasked samples captured on the first day as references. Then, the references are compared against bona fide \mf{BM0} samples captured on the second and third days of the same subjects (to compute genuine scores), as well as of other subjects (zero-effort imposter (ZEI) scores). Once genuine and ZEI comparison scores are obtained, the operating threshold is computed \pr{using the} $\tau_{FMR@0.01}$ threshold. Finally, the probe samples of \mf{each type of PA} \pr{were} compared against the reference of the same subjects separately.
In the second setting \mf{BM0-BM1}, the difference is that bona fide \mf{BM1} data captured on the second and third days are used \pr{for comparison} against references \mf{BM0} and then obtain the corresponding genuine and ZEI scores. In the third setting, BM1-BM1, the bona fide masked faces captured on the first day are references for each subject. Such references are also compared against the masked bona fide samples captured on the second and third days to obtain their genuine and ZEI scores. 
\mf{These three experimental settings are provided to enable addressing the following four questions: 1) When having an unmasked reference and we use a decision threshold that does not consider masked comparisons (BM0-BM0), how vulnerable are FR systems to the three types of attacks in CRMA (AM0, AM1, and AM2)? 
2) When having an unmasked reference and we use a decision threshold based on unmasked-to-masked comparisons (BM0-BM1), how vulnerable are FR systems to the three types of attacks in CRMA (AM0, AM1, and AM2)?  
3) When having a masked reference and we use a decision threshold based on masked-to-masked comparisons (BM1-BM1), how vulnerable are FR systems to the three types of attacks in CRMA (AM0, AM1, and AM2)? 
Additionally, we address \pr{the} fourth question: 4) will the vulnerability of FR systems be different when facing the AM1 and AM2 attacks?}
\begin{table*}[htb!]
\centering
\footnotesize
\def\arraystretch{1.2}
\resizebox{0.9\textwidth}{!}{
\begin{tabular}{|c|c|c|c|c|c|c|c|c|c|c|}
\hline
\multirow{2}{*}{Settings} & \multirow{2}{*}{Attack Probes} & \multicolumn{3}{c|}{ArcFace\cite{arcface}} & \multicolumn{3}{c|}{SphereFace \cite{sphereface}} & \multicolumn{3}{c|}{VGGFace \cite{vggface2}} \\ \cline{3-11} 
 &  & EER & GMR & IAPMR & EER & GMR & IAPMR & EER & GMR & IAPMR \\ \hline
\multirow{3}{*}{\mf{BM0 - BM0}} & AM0 & \multirow{3}{*}{0.00} & \multirow{3}{*}{100} & 98.40 {[}98.22, 98.56{]} & \multirow{3}{*}{8.57} & \multirow{3}{*}{75.85} & 66.31 {[}65.69, 66.93{]} & \multirow{3}{*}{0.12} & \multirow{3}{*}{100} & 99.47 {[}99.37, 99.56{]} \\ \cline{2-2} \cline{5-5} \cline{8-8} \cline{11-11} 
 & AM1 &  &  & 81.61 {[}81.24, 81.97{]} &  &  & 2.80 {[}2.65, 2.96{]} &  &  & 71.54 {[}71.12, 71.96{]} \\ \cline{2-2} \cline{5-5} \cline{8-8} \cline{11-11} 
 & AM2 &  &  & 97.10 {[}96.77, 97.41{]} &  &  & 10.45 {[}9.89, 11.03{]} &  &  & 97.23 {[}96.91, 97.53{]} \\ \hline
\multirow{3}{*}{\mf{BM0 - BM1}} & AM0 & \multirow{3}{*}{2.25} & \multirow{3}{*}{96.56} & 98.73 {[}98.58, 98.88{]} & \multirow{3}{*}{22.83} & \multirow{3}{*}{19.99} & 84.17 {[}83.68, 84.64{]} & \multirow{3}{*}{2.29} & \multirow{3}{*}{94.2} & 99.86 {[}99.80, 99.90{]} \\ \cline{2-2} \cline{5-5} \cline{8-8} \cline{11-11} 
 & AM1 &  &  & 88.57 {[}88.27, 88.86{]} &  &  & 15.26 {[}14.92, 15.60{]} &  &  & 90.24 {[}89.96, 90.51{]} \\ \cline{2-2} \cline{5-5} \cline{8-8} \cline{11-11} 
 & AM2 &  &  & 98.56 {[}98.33, 98.78{]} &  &  & 40.00 {[}39.09, 40.91{]} &  &  & 99.55 {[}99.41, 99.67{]} \\ \hline
\multirow{3}{*}{\mf{BM1 - BM1}} & AM0 & \multirow{3}{*}{1.00} & \multirow{3}{*}{99.00} & 70.62 {[}70.19, 71.04{]} & \multirow{3}{*}{13.13} & \multirow{3}{*}{59.33} & 2.43 {[}2.29, 2.58{]} & \multirow{3}{*}{0.85} & \multirow{3}{*}{99.46} & 45.84 {[}45.38, 46.31{]} \\ \cline{2-2} \cline{5-5} \cline{8-8} \cline{11-11} 
 & AM1 &  &  & 94.20 {[}94.04, 94.35{]} &  &  & 47.69 {[}47.36, 48.02{]} &  &  & 97.41 {[}97.30, 97.51{]} \\ \cline{2-2} \cline{5-5} \cline{8-8} \cline{11-11} 
 & AM2 &  &  & 97.70 {[}97.49, 97.89{]} &  &  & 50.82 {[}50.16, 51.48{]} &  &  & 98.26 {[}98.08, 98.43{]} \\ \hline
\end{tabular}}
\caption{The performance and vulnerability of FR systems. The GMR and IAPMR values \pr{were} computed based on the $\tau_{FMR@0.01}$ threshold. The 95\% confidence intervals for the IAPMR values are shown in \pr{parentheses}.}
\vspace{-4mm}
\label{tab:iapmr}
\end{table*}

\subsection{Evaluation metrics}
To measure the performance of FR techniques, the \textit{genuine match rate} (GMR), \pr{which refers} to the proportion of correctly matched genuine samples, is used at \mf{the} fixed false match rate (FMR). GMR is equal to 1 minus the false non-match rate (FNMR). Moreover, to analyze the vulnerability of FR algorithms for our masked attacks, \pr{the} \textit{imposter attack presentation match rate} (IAPMR) corresponding to the proportion of PAs accepted by the FR system as genuine presentations is adopted. IAPMR also follows the standard definition presented in ISO/IEC 30107-3 \cite{ISO301073}. The threshold for GMR and IAMPR is defined by fixing the FMR at 1\% (denoted as $\tau_{FMR@0.01}$). The probe images with similarity scores lower than the $\tau_{FMR@0.01}$ are not matched. Moreover, the recognition score-distribution histograms are shown in Fig.~\ref{fig:arcface},~\ref{fig:sphereface}, and \ref{fig:vggface}. \pr{In addition to} these metrics, the EER value, where FMR equals FNMR, is computed to compare the FR algorithms.

\subsection{Analysis of FR results}

\begin{figure}[htb!]
\begin{center}
\includegraphics[width=0.8\linewidth]{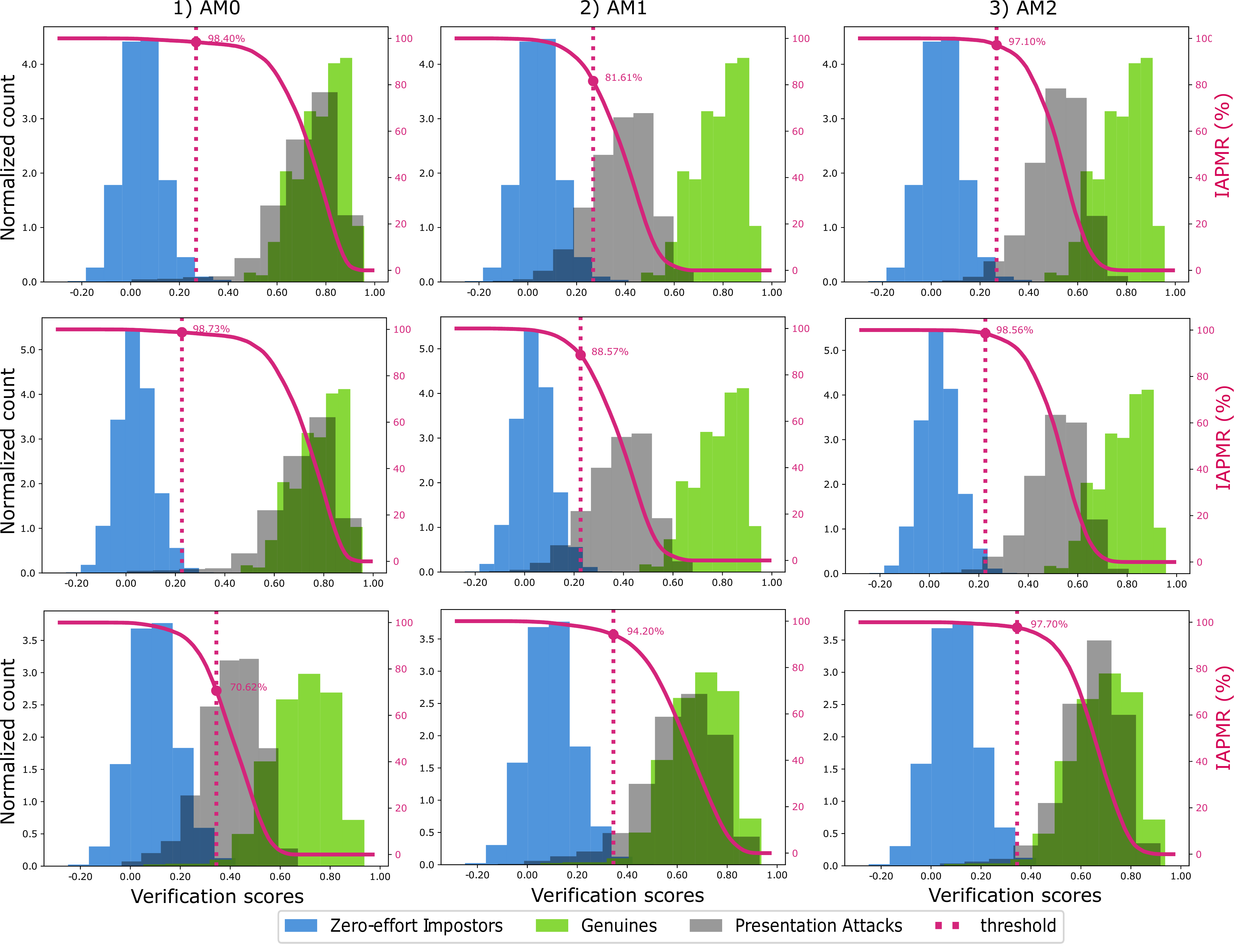}
\end{center}
\caption{The similarity score distributions by off-the-shelf ArcFace \cite{arcface}. The rows from top to bottom represent three experimental settings: \mf{BM0-BM0, BM0-BM1, BM1-BM1}, as shown in Tab.~\ref{tab:iapmr}.}
\label{fig:arcface}
\end{figure}

\begin{figure}[htb!]
\begin{center}
\includegraphics[width=0.8\linewidth]{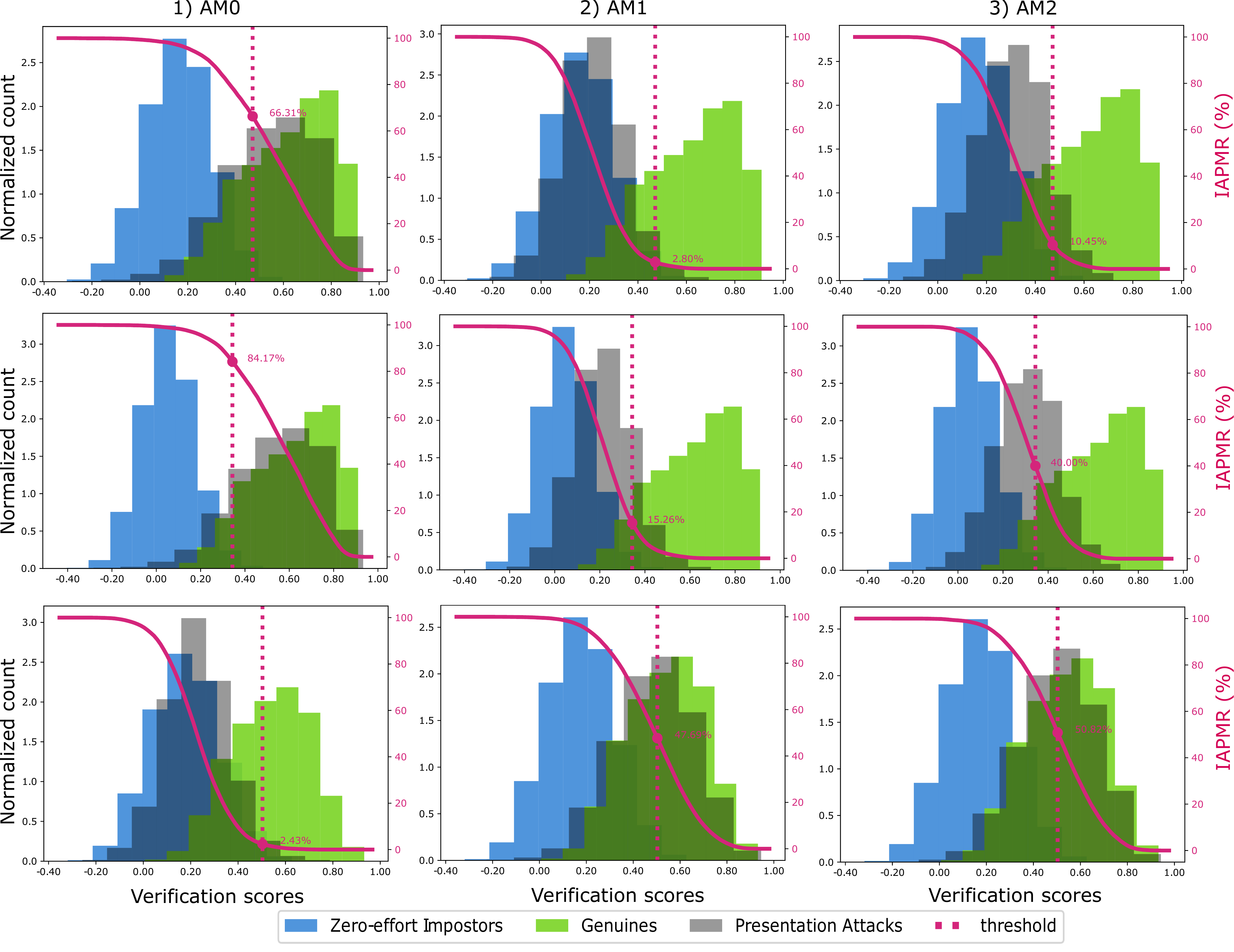}
\end{center}
\caption{The similarity score distributions by off-the-shelf SphereFace \cite{sphereface}.}
\label{fig:sphereface}
\end{figure}

\begin{figure}[htb!]
\begin{center}
\includegraphics[width=0.8\linewidth]{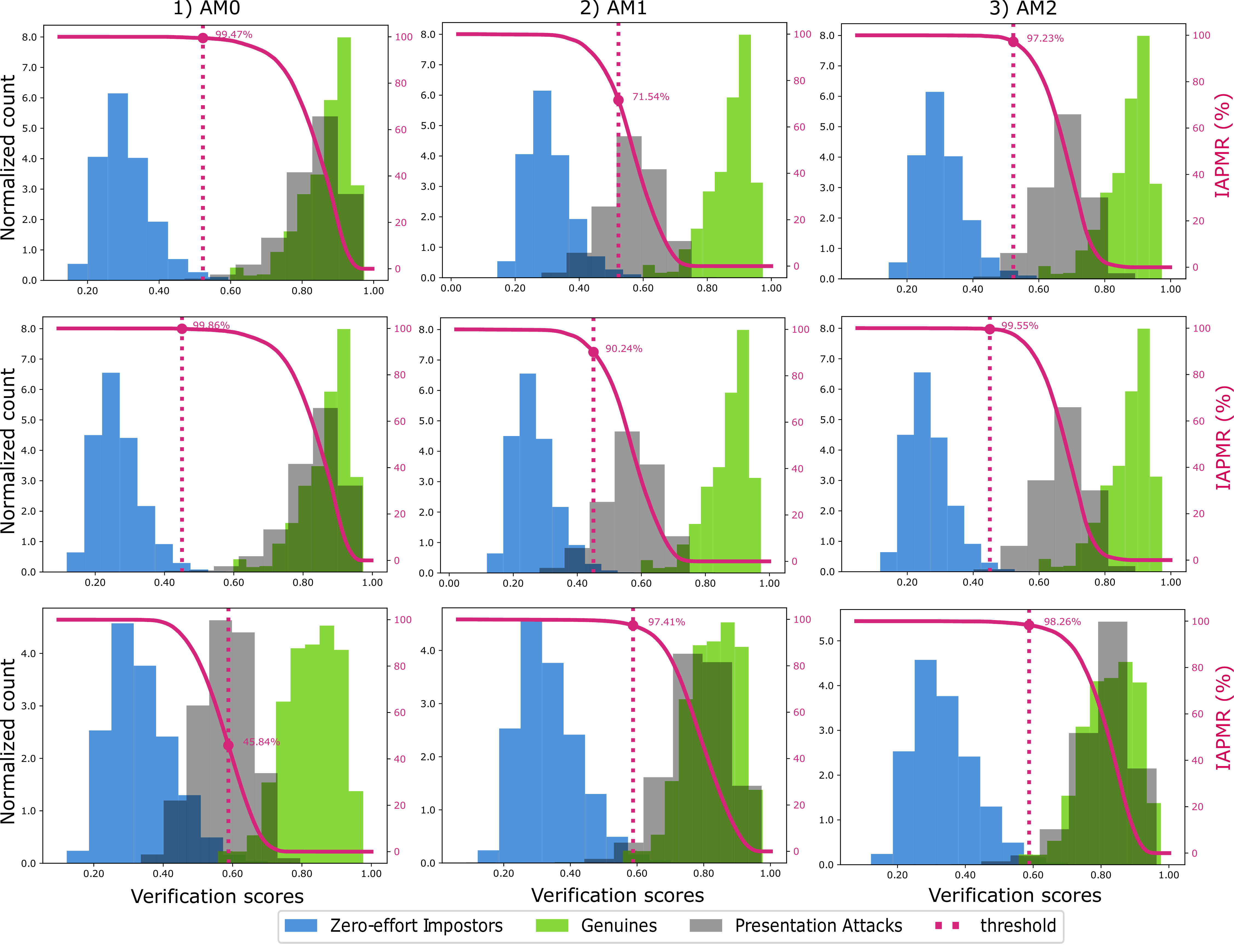}
\end{center}
\caption{The similarity score distributions by off-the-shelf VGGFace \cite{sphereface}.}
\vspace{-4mm}
\label{fig:vggface}
\end{figure}

The performance and vulnerability of each FR system \pr{are} summarized in Tab.~\ref{tab:iapmr}. SphereFace \cite{sphereface} obtains relatively low IAPMR values; however, its GMR values are also much lower than \pr{those of} ArcFace \cite{arcface} and VGGFace \cite{vggface2}. In general, the IAPMR values of all three FR systems \pr{were} close to their GMR values. Specifically, FR systems are vulnerable to unmasked attacks when unmasked bona fide samples are used as references (the settings BM0-BM0 and BM0-BM1), and vulnerable to the masked attacks when \pr{the} reference is masked bona fide data. Comparing the vulnerability analysis results \pr{for} AM1 and AM2 in all three cases and all FR systems, we note that the IAMPR values of AM2 are always significantly higher than \pr{those} of AM1. This indicates that applying real masks on attack presentations can further reduce the performance of FR systems. This might be due to the fact that the AM2 attacks possess more realistic features than AM1. 
To further verify this assumption, we provide histograms of the similarity score distribution in the three scenarios and three FR systems (see Fig.~\ref{fig:arcface},~\ref{fig:sphereface}, and \ref{fig:vggface}). In the histograms, green refers to genuine scores, blue \pr{represents} ZEI scores, and \pr{gray represents} attack verification scores. The ideal situation is that \pr{there is} no overlap between the green and the other two histograms. Fig.~\ref{fig:arcface} shows the score distributions of ArcFace \cite{arcface}, where the rows from top to bottom represent \mf{BM0-BM0, BM0-BM1, BM1-BM1} cases and columns from left to right refer to AM0, AM1, and AM2 attacks. It can be seen that 1) the verification scores of attacks are higher than \pr{the} scores of ZEI in all cases. 2) The scores of AM0 attacks and genuine scores almost overlap in the \mf{BM0-BM0 and BM0-BM1} settings, while the scores of AM1/AM2 attacks have \pr{many} overlapping areas with genuine scores in the \mf{BM1-BM1} setting. 3) for all cases, the scores of AM2 \pr{have} more overlaps with genuine scores than AM1. Similar observations can be found in Fig.~\ref{fig:sphereface} for \pr{the} SphereFace, and Fig.~\ref{fig:vggface} for VGGFace. These observations are consistent with the findings \pr{presented} in Tab.~\ref{tab:iapmr}.

\mf{Overall, these results indicate that 1) FR systems are more vulnerable to unmasked attacks compare to masked attacks when the references are unmasked faces, 2) when the threshold is computed based on the unmasked-to-masked comparison, the vulnerability of FR systems becomes higher for both masked or unmasked attacks, 3) when the reference is masked, FR systems are more vulnerable to masked attacks in comparison to the FR systems having unmasked references. Another important finding is that 4) FR systems pose a higher vulnerability for spoof faces with real masks placed on them (AM2) than a masked face attack (AM1). Such observations raise concerns about the security of FR systems when facing masked attacks.}

\vspace{-3mm}
\section{Conclusion}
\label{sec:conclusion}

\pr{We studied} the behavior of PAD methods \pr{on} different types of masked face images.
To enable our study, we presented a new large-scale face PAD database, CRMA, including the conventional unmasked attacks, novel attacks with faces wearing masks, and attacks with real masks placed on spoof faces. It consists of 13,113 high-resolution videos and has a large diversity in capture sensors, displays, and capture scales. To study the effect of wearing a mask on \pr{the} PAD algorithms, we designed three experimental protocols. The first protocol measures the generalizability of the current PAD algorithms on unknown masked bona fide or attack samples. \pr{In} the second protocol, masked data \pr{are} used in the training phase to measure the performance of PAD solutions where the face masks are known. The third protocol investigates the generalizability of models trained on masked face attacks when tested on attacks covered by a real mask. Extensive experiments were conducted \pr{using} these protocols. The results showed that PAD algorithms have a high possibility of detecting masked bona fide samples as attackers (median BPCER value for BM1 in protocol-1 is 48.25\%). Moreover, even if the PAD solutions have seen the masked bona fide data, the PAD algorithms still perform worse on masked bona fide samples \pr{compared with} unmasked bona fides. Furthermore, the PAD solutions trained on masked face attacks (AM1) do not generalize well on attacks covered by a real mask (AM2). For example, the APCER values achieved by DeepPixBis increased from 0.62\% for AM1 to 2.92\% for AM2 in print attack and from 1.92\% for AM1 to 6.43\%  for AM2 in replay  (protocol-3). 
\pr{In addition,} we performed \pr{a} thorough \pr{analysis} of the vulnerability of FR systems to such novel attacks. The results indicate that FR systems are vulnerable to both masked \pr{and} unmasked attacks. For example, when the reference images and system threshold are based on unmasked faces (BM0-BM0), the IAPMR values for unmasked attacks (AM0), masked attacks (AM1), and attacks covered by a real mask (AM2) are 98.40\%, 81.60\%, and 97.10\%, respectively. This leads to the interesting observation that all the investigated FR systems are more vulnerable to attacks where real masks are placed on attacks (AM2) than attacks of masked faces (AM1). 

\noindent\textbf{Acknowledgments:}
This research work has been funded by the German Federal Ministry of Education and Research and the Hessen State Ministry for Higher Education, Research and the Arts within their joint support of the National Research Center for Applied Cybersecurity ATHENE.

\vspace{-3mm}

{\small
\bibliographystyle{IEEEtran}
\bibliography{mybibfile}
}

\newpage
\clearpage

\textbf{Supplementary material}


\section{Visualization and analysis}
\begin{figure}[htb!]
\begin{center}
\includegraphics[width=0.99\linewidth]{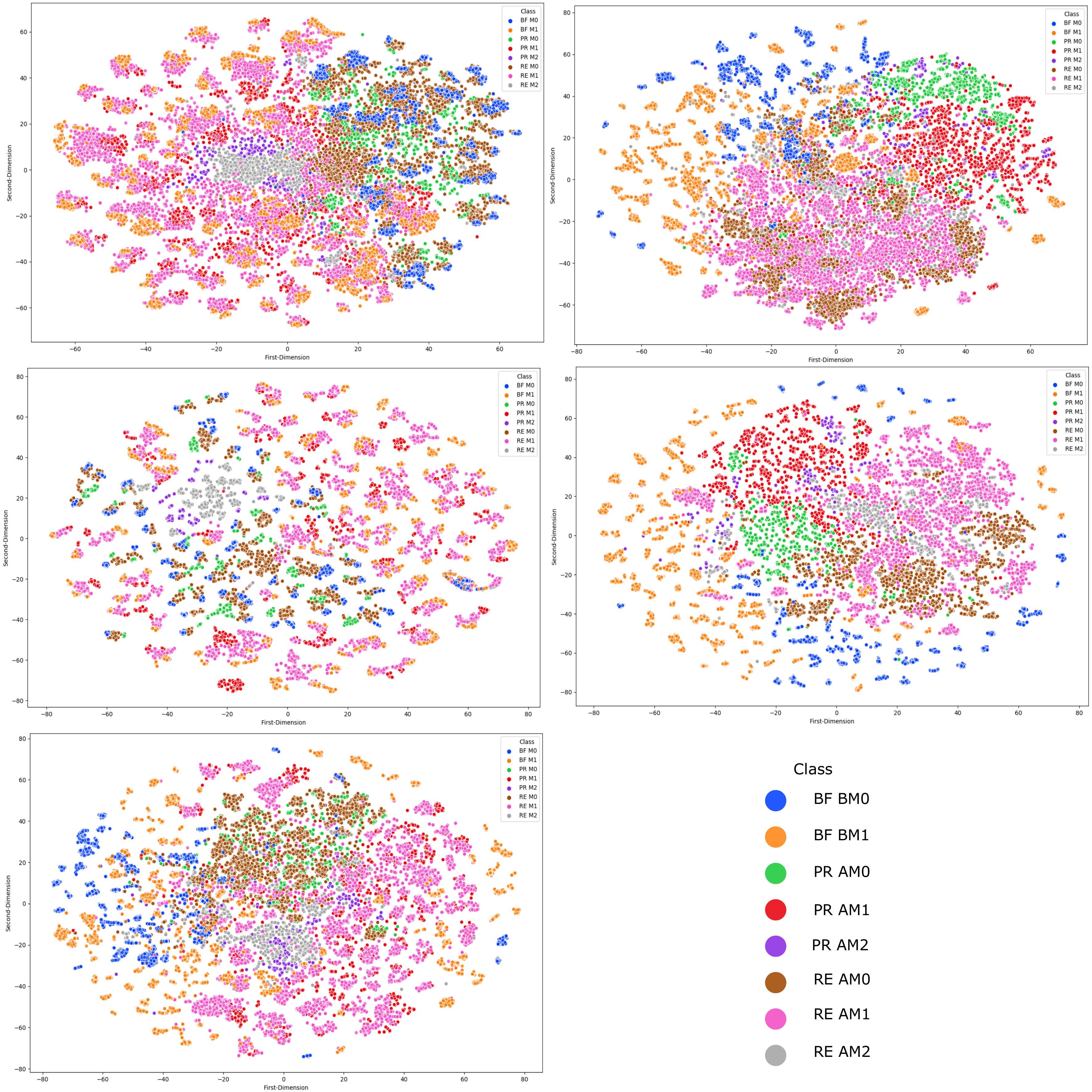}
\end{center}
\caption{The t-SNE plots of fine-tuned $\mathrm{Inception_{FT}}$ and $\mathrm{FASNet_{FT}}$ (in left column), and trained from scratch $\mathrm{Inception_{TFS}}$ and $\mathrm{FASNet_{TFS}}$ (in right column) on the first protocol which targets the unknown masked bona fide and two types of masked attacks. These plots show that fine-tuned $\mathrm{Inception_{FT}}$ and $\mathrm{FASNet_{FT}}$ cannot discriminate the features between bona fide and attacks. The 2D features representing the face images seem to be grouped based on the existence of face masks. Moreover, AM2 attacks are surrounded by bona fide samples and other types of attacks.}
\vspace{-5mm}
\label{fig:feature_tsne_p1}
\end{figure}

\begin{figure}[htb!]
\begin{center}
\includegraphics[width=0.99\linewidth]{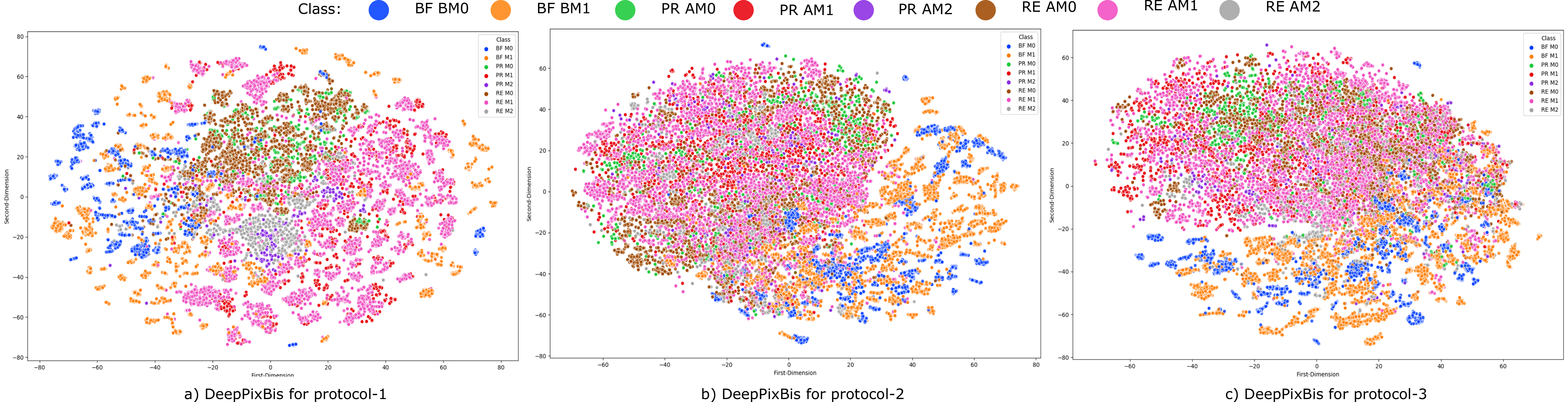}
\end{center}
\caption{The t-SNE plots for DeepPixBis method on three protocols in our CRMA database. It can be seen that the unknown \mf{AM2} attacks are enclosed by bona fide samples and other types of attacks.}
\vspace{-5mm}
\label{fig:deeppix_tsne}
\end{figure}

To further observe and deeper understand the discriminative features between bona fide and PAs in the CRMA database, we present here, as supplementary material, the visualized features of face samples from seven classes: \mf{BF-BM0, BF-BM1, PR-AM0, PR-AM1, PR-AM2, RE-AM0, RE-AM1, RE-AM2}. Here, BF refers to bona fide, while PR and RE correspond to print and reply. Fig.~\ref{fig:feature_tsne_p1} shows the differential results of fine-tuned $\mathrm{Inception_{FT}}$ and $\mathrm{FASNet_{FT}}$, and trained from scratch $\mathrm{Inception_{TFS}}$ and $\mathrm{FASNet_{TFS}}$ on the first protocol. Furthermore, the results of DeepPixBis on three experimental protocols are selected to present in Fig.~\ref{fig:deeppix_tsne} as the DeepPixBis method outperforms other PAD algorithms on CRMA database.

First, by observing the t-SNE plots from $\mathrm{Inception_{FT}}$ and $\mathrm{FASNet_{FT}}$ methods in Fig.~\ref{fig:feature_tsne_p1}, we \mf{observed that BM0, AM0 of print and replay PAIs are tightly grouped}, while \mf{masked BM1 and AM1} samples in both PAIs are clustered. Simultaneously, the print \mf{AM2} and replay \mf{AM2} attack samples tend to form a compact distribution. Moreover, it is interesting to note that \mf{AM2} attacks are enclosed by the previous \mf{BM0/AM0} and \mf{BM1/AM1} clusters. This indicated that 1) fine-tuned networks are unable to learn the discriminative features between bona fide and attacks. 2) \mf{AM2} attacks, which have artifacts and live features together, are more difficult for networks to make a correct PAD decision. Second, it can be seen in the t-SNE subplots from $\mathrm{Inception_{TFS}}$ and $\mathrm{FASNet_{TFS}}$ that compared to fine-tuned networks, trained from scratch networks perform better. Because the features of \mf{BM0 and BM1} are in a group, print attacks in a group, and replay attacks in a group. However, 2D features of \mf{AM2} attacks in print and replay PAIs are still closer than other attack types when employing $\mathrm{FASNet_{TFS}}$. Overall, \mf{AM2} attacks are slightly difficult to detected correctly than \mf{AM1} attacks by trained from scratch networks, even though the bona fide and attacks are more separate than using fine-tuned networks. Third, by taking a look at results on \mf{protocol-1} by DeepPixBis method in Fig.~\ref{fig:deeppix_tsne}.(a), almost half of the \mf{BM1} data is mixed with attack data. Besides, on the \mf{protocol-1} plot, print \mf{AM0} data is close to the replay \mf{AM0}, print \mf{AM1} is close to the replay \mf{AM1}, and grouped \mf{AM2} attacks in both PAIs. It should be noticed again that \mf{AM2} attacks are surrounded by the bona fide and other attack data. In Fig.~\ref{fig:deeppix_tsne}.(b) and Fig.~\ref{fig:deeppix_tsne}.(c), bona fide and attack data are more separate. However, some replay \mf{AM2} attacks are still mixed inside bona fide groups on the \mf{protocol-2}. Additionally, more \mf{AM2} attacks are close to the bona fide samples on the \mf{protocol-3 than protocol-2}. A possible reason is that \mf{AM2} attacks are not learned in the training phase of \mf{protocol-3}. All the above findings are consistent with observations in our main work.

Together these results provide important insights that 1) fine-tuned networks have poorer generalizability on the unknown masked bona fide and attack data than trained from scratch networks. 2) masked bona fide samples are more probable to detected as bona fide by the pre-COVID-19 PAD algorithms. 3) \mf{attacks with real masks on spoofing faces (AM2) are more accessible detected by PAD systems as bona fide than attacks with masked faces (AM1).}   

\subsection{The Cross-database evaluation}
\label{ssec:cross_database}

\begin{table*}[htb!]
\footnotesize
\centering
\resizebox{0.9\textwidth}{!}{
\begin{tabular}{|c|c|c|c|c|c|c|c|}
\hline
\multirow{3}{*}{Train} & \multirow{3}{*}{Method} & \multicolumn{6}{c|}{Threshold @ BPCER 10\% in dev set of CRMA database} \\ \cline{3-8} 
 &  & \multicolumn{2}{c|}{CASIA-FASD (\%)} & \multicolumn{2}{c|}{MSU-MFSD (\%) } & \multicolumn{2}{c|}{Oulu-NPU (\%)} \\ \cline{3-8} 
 &  & D-EER & HTER & D-EER & HTER & D-EER & HTER \\ \hline
\multirow{5}{*}{P1} & LBP & 41.25 & 47.19 & 41.67 & 40.28 & 32.50 & 38.96 \\ \cline{2-8} 
 & $\mathrm{Inception_{FT}}$ & 37.19 & 42.19 & 36.81 & 34.03 & 24.90 & 24.06 \\ \cline{2-8} 
 & CPqD & \textbf{36.25} & 46.56 & 34.03 & 31.94 & 22.50 & 23.23 \\ \cline{2-8}
 & $\mathrm{FASNet_{FT}}$ & 47.50 & 56.56 & 40.97 & 44.44 & 18.33 & 18.33 \\ \cline{2-8} 
 & $\mathrm{Inception_{TFS}}$ & 40.00 & 51.88 & 36.81 & \textbf{29.17} & \textbf{9.17} & 17.40 \\ \cline{2-8}
 & $\mathrm{FASNet_{TFS}}$ & 48.44 & \textbf{41.88} & \textbf{20.83} & 36.81 & \textbf{9.17} & \textbf{13.12} \\ \cline{2-8}
 & DeepPixBis & 47.50 & 49.06 & 41.67 & 39.58 & 21.56 & 24.23 \\ \hline \hline
\multirow{5}{*}{P2} & LBP & 46.25 & 45.63 & 45.83 & 44.44 & 30.83 & 30.83 \\ \cline{2-8} 
 & $\mathrm{Inception_{FT}}$ & \textbf{34.69} & \textbf{34.69} & 45.83 & 42.36 & 20.83 & 37.19 \\ \cline{2-8} 
 & CPqD & 40.00 & 45.63 & 45.14 & 47.92 & 21.67 & 29.27 \\ \cline{2-8} 
 & $\mathrm{FASNet_{FT}}$ & 50.00 & 55.94 & 40.97 & 39.58 & 19.27 & 26.77 \\ \cline{2-8} 
 & $\mathrm{Inception_{TFS}}$ & 37.50 & 47.50 & \textbf{25.00} & \textbf{30.56} & \textbf{13.96} & \textbf{14.48} \\ \cline{2-8}
 & $\mathrm{FASNet_{TFS}}$ & 47.19 & 44.38 & 37.50 & 31.25 & 16.67 & 22.81 \\ \cline{2-8}
 & DeepPixBis & 46.25 & 46.88 & 45.83 & 38.89 & 15.94 & 28.96 \\ \hline \hline
\multirow{5}{*}{P3} & LBP & 47.50 & 46.56 & 45.83 & 44.44 & 31.67 & 32.08 \\ \cline{2-8} 
 & $\mathrm{Inception_{FT}}$ & \textbf{37.50} & \textbf{33.44} & 41.67 & 40.28 & 20.10 & 33.96 \\ \cline{2-8} 
 & CPqD & 43.75 & 47.50 & 45.83 & 43.75 & 21.46 & 28.12 \\ \cline{2-8} 
 & $\mathrm{FASNet_{FT}}$ & 53.75 & 55.31 & 41.67 & 40.97 & 15.94 & \textbf{17.92} \\ \cline{2-8}
 & $\mathrm{Inception_{TFS}}$ & 42.50 & 55.62 & \textbf{25.00} & 30.56 & 15.00 & 40.73 \\ \cline{2-8}
 & $\mathrm{FASNet_{TFS}}$ & 46.69 & 49.38 & 28.47 & \textbf{27.78} & \textbf{13.96} & 23.75 \\ \cline{2-8}
 & DeepPixBis & 43.75 & 42.81 & 42.36 & 36.81 & 16.46 & 28.65 \\ \hline
\end{tabular}}
\caption{Cross-database evaluation 2: trained on different protocols of the CRMA database and tested on three public databases. \mf{The bold numbers refer to the lowest error rates of all PAD methods in each dataset and each protocol.}}
\vspace{-5mm}
\label{tab:cross_db_2}
\end{table*}

\mf{The first cross-database scenario is similar to protocol-1 and discussed in the main work. The second cross-database scenario can show the diversity of our CRMA database due to the face masks, various sensors,  and different scales, as the results are reported in the Tab.~\ref{tab:cross_db_2}. It can be seen that trained from scratch networks outperform traditional LBP or fine-tuned networks in most cases. The models trained on protocol-1 of the CRMA database achieved better results than models trained on protocol-2 and protocol-3 in the OULU-NPU database (13.12\% HTER on protocol-1, 14.48\%, and 17.92\% on protocol-2 and protocol-3). We can conclude that even without masked data, the CRMA database still possesses great diversity in sensors and environments to boost the performance of vanilla models. For example, $\mathrm{FASNet_{TFS}}$, which train VGG16 from scratch, achieves competitive results (13.12\% HTER) on OULU-NPU database \cite{pad_competition, oulu_npu}. Besides, as mentioned earlier, even though DeepPixBis achieves much better results than other methods in intra-database evaluations, it performs mostly worse than other deep-learning methods, especially cross-testing on MSU-MFSD database.}

\end{document}